\documentclass[conference]{IEEEtran}
\IEEEoverridecommandlockouts
\usepackage{cite}
\usepackage{amsmath,amssymb,amsfonts}
\usepackage{algorithmic}
\usepackage{graphicx}
\usepackage{multirow}
\usepackage{multicol}
\usepackage{textcomp}
\usepackage{xcolor}
\usepackage{graphicx}
\ifCLASSOPTIONcompsoc
\usepackage[caption=false,font=normalsize,labelfon
t=sf,textfont=sf]{subfig}
\else
\usepackage[caption=false,font=footnotesize]{subfig}
\fi

\def\BibTeX{{\rm B\kern-.05em{\sc i\kern-.025em b}\kern-.08em
    T\kern-.1667em\lower.7ex\hbox{E}\kern-.125emX}}
\begin{document}

\title{Where am I?  SLAM for Mobile Machines \\ on A Smart Working Site\\
}
\makeatletter
\newcommand{\linebreakand}{%
  \end{@IEEEauthorhalign}
  \hfill\mbox{}\par
  \mbox{}\hfill\begin{@IEEEauthorhalign}
}

\makeatother

\author{
  \IEEEauthorblockN{1\textsuperscript{st} Yusheng Xiang, 7\textsuperscript{th} Marcus Geimer}
  \IEEEauthorblockA{\textit{Institute of Vehicle System Technology} \\
    \textit{Karlsruhe Institute of Technology}\\
    Karlsruhe, Germany \\
    yusheng.xiang@partner.kit.edu}
  \and
  \IEEEauthorblockN{2\textsuperscript{nd} Dianzhao Li }
  \IEEEauthorblockA{\textit{Institute of Vehicle System Technology} \\
    \textit{Karlsruhe Institute of Technology}\\
    Karlsruhe, Germany \\
    dianzhao.li@kit.edu}
  \and
  \IEEEauthorblockN{3\textsuperscript{rd} Tianqing Su}
  \IEEEauthorblockA{\textit{Guanghua School of Management} \\
    \textit{Peking University}\\
    Beijing, China \\
    tianqing.su@hotmail.com}
  \linebreakand 

  \IEEEauthorblockN{4\textsuperscript{th} Quan Zhou}
  \IEEEauthorblockA{\textit{Vehicle and Engine Research Centre} \\
    \textit{University of Birmingham}\\
    Birmingham, UK \\
    q.zhou@bham.ac.uk}
  \and
  \IEEEauthorblockN{5\textsuperscript{th} Christine Brach}
      \IEEEauthorblockA{\textit{Division of Mobile Hydraulics} \\
    \textit{Robert Bosch GmbH}\\
    Elchingen, Germany \\
    christine.brach@boschrexroth.de}
  \and
  \IEEEauthorblockN{6\textsuperscript{th} Samuel S.  Mao}
  \IEEEauthorblockA{\textit{Department of Mechanical Engineering} \\
    \textit{University of California, Berkeley}\\
    Berkeley, USA \\
    ssmao@berkeley.edu}

}
\maketitle

\begin{abstract}

The current optimization approaches of construction machinery are mainly based on internal sensors. However, the decision of a reasonable strategy is not only determined by its intrinsic signals, but also very strongly by environmental information, especially the terrain. Due to the dynamically changing of the construction site and the consequent absence of a high definition map, the Simultaneous Localization and Mapping (SLAM) offering the terrain information for construction machines is still challenging. Current SLAM technologies proposed for mobile machines are strongly dependent on costly or computationally expensive sensors, such as RTK GPS and cameras, so that commercial use is rare. In this study, we proposed an affordable SLAM method to create a multi-layer gird map for the construction site so that the machine can have the environmental information and be optimized accordingly. Concretely, after the machine passes by, we can get the local information and record it. Combining with positioning technology, we then create a map of the interesting places of the construction site. As a result of our research gathered from Gazebo, we showed that a suitable layout is the combination of 1 IMU and 2 differential GPS antennas using the unscented Kalman filter, which keeps the average distance error lower than 2m and the mapping error lower than 1.3\% in the harsh environment. As an outlook, our SLAM technology provides the cornerstone to activate many efficiency improvement approaches.

\end{abstract}

\begin{IEEEkeywords}
Unscented Kalman filter, Localization of construction machines, Smart working site, SLAM, ROS
\end{IEEEkeywords}

\section{Introduction}

 \begin{figure}
     \centering
    \includegraphics[width=3.5in]{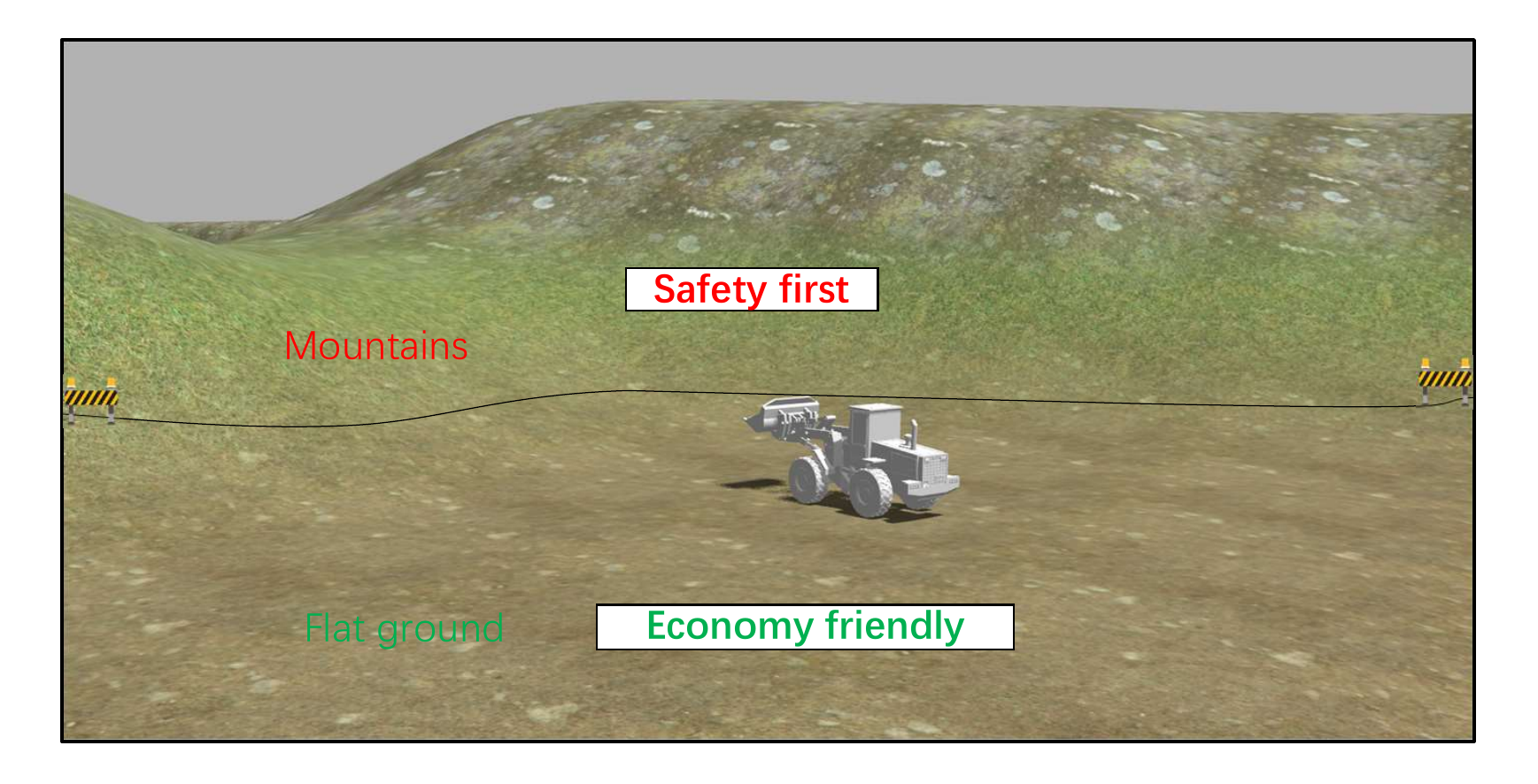}
   \caption{Mobile machines perform tasks more efficiently or safer according to their location and surroundings information. The short-term goal of SLAM is to prevent construction machinery from always working in low-efficiency areas for safety reasons, whereas the long-term goal is to increase the productivity of the working site with the help of path planning. The study focuses on affordable SLAM technology for construction machines.}
     \label{fig:wheel loader in working site}
  \end{figure}

Over the past several decades, the operation strategies aiming to increase mobile machines' holistic efficiency receive much attention. As Osinenko pointed out, the engine power of farm tractors is growing at 1.8 kW per year and has reached about 500KW for the most powerful traction class \cite{Osinenko.2015}, indicating the great potential to increase the holistic efficiency of the machines. Current solutions to optimize the system can be concluded as two kinds. To overcome the extreme conditions, the basic idea of the first method is equipping the mobile machines with an energy-saving system \cite{Sprengel.2015, Vukovic.2013, Vukovic.2016}, such as accumulators. The other one is based on a dramatic controller to increase the dispensable power in a short period \cite{ Xiang.2020d, Ge.2017, Ge.2018, Joos.2020,Xiang.2020c, Mutschler.2018}. However, due to safety reasons, most of the operation strategies of mobile machines are designed conservatively. For instance, the energy-saving equipment should always at a relatively high level, and the controller must be designed to have rapid responses. As a result of that, the machines are still very likely to work in suboptimal areas. In the previous studies, most of them utilize the intrinsic signals to optimize the system \cite{Xiang.2020b} since the environmental information is hardware or computationally expensive to be gathered. However, the environment also has an essential influence on the system, i.e., to perform tasks both efficiently and safely, the construction machines shall be optimized by knowing their location and surroundings; thus, we proposed a method that can generate the map information surrounding the mobile machines only with affordable sensors so that provides the possibility to improve the system further. The basic idea of our approach is to generate the map information of the working site based on the vehicle position, rolling resistance, as well as road grade. Concretely, a special recursive least square with forgetting algorithm is used to record the road grade and the rolling resistance in realtime \cite{Vahidi.2005, Xiang.2020}. These information will be saved together with the localization information. Consequently, after the machine passes by, it will record the information about that place. Since the mobile machines are driving repeatably for a special task, the method can be expected to work well even when the map information does not cover most of the working site. Fig. \ref{fig:wheel loader in working site} illustrates the motivation of our approach.

\section{Problem Statement}

Although the map information can also be obtained from satellite, it is impossible to get the valuable information, such as a High Definition (HD) map, only depending on remote sensing due to a construction site's fast-changing environment. Also, the sensors can be quite noisy. Especially, they will be further exacerbated on a working site. The sensors, such as Global Positioning System (GPS), Inertial Measurement Unit (IMU),  and odometry sensor will have higher measurement errors in case of  the harsh environment. In addition, different construction machines have different drivetrain system, which makes a predefined motion model difficult. For instance, since mobile machines may work outside the coverage of base stations, the GPS signal can only achieve nearly 10 m accuracy \cite{Tan.2006} due to the lack of error correction. Last but not least, for passenger cars, the longitude error might not have such a negative effect as the latitude error since further measurements can be adopted to avoid the collision. In the case of construction machines, both errors shall be treated equally.  

\section{Goal of this research}
The goals of this paper are twofold. 
The first goal of the paper is to find out the most suitable sensor arrangements for construction machines. For accurate estimation of the position of machines, rather than only trust the measurement from one sensor, we fuse a series of different kinds of sensors with the aim of sensor fusion technology, derived from Kalman filter \cite{Kalman.1960}, to achieve better accuracy. Afterward, the second goal is to create a map with the environment condition by combining the surface resistance, road grade, and position information in realtime. Thanks to this map, further optimization of operation strategy and path planning can be realized. Although we measure the surface resistance and road grade by recursive least square with multi forgetting factors, the map-building approach we proposed can also be combined with other methods with other kinds of sensors, such as using ultrasonic proposed by Jung \cite{Jung.2009}.

\section{Related works}

\subsection{Sensors}

The combination of several sensing systems so that they can compensate the technical shortcoming of each other is well-known in the field of autonomous systems \cite{Wang.2016, Perea.2013, Sukkarieh.1999}. Therefore, there are a series of researches focusing on sensor fusion. Here we first summary the commonly used sensors for simultaneous localization and mapping (SLAM). Although we agree the introduction of the HD map can surely increase the accuracy of the localization, we do not consider this technology for the construction machines due to the dynamically changing of the construction site, as mentioned in  \cite{Iwataki.2015}. 

\subsubsection{GPS}

Global navigation satellite systems (GNSS) such as GPS, GLONASS, BeiDou, and Galileo rely on at least four satellites to estimate global position at a relatively low cost. Typical standalone GPS average accuracy ranges from few meters to above 20 m \cite{Tan.2006} due to ionospheric delay, multipath effects, ephemetrics $ \& $ clock errors, and Geometric dilution of precision (GDOP).
To improve the accuracy, one of the most used techniques is Differential GPS (DGPS), which utilizes measurements from an onboard vehicle GPS unit and a GPS unit on a fixed infrastructure unit with a known location. Here the known fixed infrastructure unit is called Reference Station, which calculates the local error in the GPS position measurement periodically. The onboard vehicle GPS units then use this correction to adjust their own GPS estimation. According to \cite{Skog.2009, Isa.2017, Kim.2017, Zhang.2020, Soatti.2018}, an average accuracy in the range of 1–2 m can be achieved, mainly depending on the distance between the vehicle and the base station. 
Another commonly used improvement is realtime kinematic (RTK) GPS, which estimates relative position by means of the phase of the carrier signal and can be expected to achieve centimeter-level accuracy. Notice that, both of them depend on a fixed base station with a known position nearby, through the principle of them is quite different. In Oct 2020, when we write this paper, RTK GPS is still an extraordinary expensive approach and usually be used to define the ground truth position of vehicles. Some low-cost RTK GPS sensors, under 1000 bulks, are designed with much lower receive frequency \cite{Skoglund.2016} and thus cause problems as vehicles driving fast. Thus, in most commercial uses, DGPS is preferable for reducing the cost. Therefore, in our research, we conservatively considered the accuracy of DGPS as 2m, which is consistent with the normal performance of DGPS. Obviously, as the performance, especially the accuracy, of the GPS increases, our mapping approach will also have better performance consequently.


\subsubsection{IMU}

Inertial measurement units (IMUs) are integrated electronic devices that contain accelerometers, magnetometers, and gyroscopes. It can provide raw IMU measurements to calculate attitude, angular rates, linear velocity, and position relative to a global reference frame. 

\subsubsection{Odometry}

Odometry is the most widely used navigation method for positioning; it provides good short-term accuracy, is inexpensive, and allows very high sampling rates. Odometry is based on simple equations, which hold true when wheel revolutions can be translated accurately into linear displacement relative to the floor. The main advantage of odometry is that all localization information comes from the vehicle itself so that this information is always available. Usually, it is the only localization information when other sensors are not able to provide data. Thus, a good odometry based localization system is always necessary, and it is usually the first step to localization \cite{Toledo.2018}.


\subsection{Localization technologies}

\subsubsection{Mobile robotics}
a series of researches using sensor fusion to achieve a highly accurate localization has been studied worldwide. The technologies about SLAM can be roughly divided into two parts: indoors and outdoors. For indoor localization, such as domestic robots \cite{Qi.2020}, a GPS system cannot be used. However, the road is relatively flat, and thus only a two-dimensional map is needed. In contrast, when it comes to offload navigation in rough terrain, the algorithms must be capable to handle three dimensions of the environment. After the success of Kalman filter \cite{Kalman.1960}, extended Kalman filter \cite{Julier.1997}, and finally unscented Kalman filter \cite{Wan.2000, Welch.1995}, the idea that a mobile robot which executes useful missions should be endowed with navigation ability has become a consensus. However, the selection of combining different sensors is from case to case different. For instance, Bento fused the data from ABS sensors and GPS for outdoor localization, based on extended Kalman filter \cite{Bento.2005}, and Zhang integrated the information from GPS and IMU \cite{Zhang.2012}. Also, Li used a camera instead of GPS to accomplish mean positioning errors of 75 cm \cite{Li.2013}. In addition, Wolcott proposed a Visual Localization method within LIDAR Maps for Automated Urban Driving \cite{Wolcott.2014}. For the cost purpose, Ward studied the possibility to use radar to localize the vehicle's position and demonstrates that errors go to 27.8 cm laterally and 115.1 cm longitudinally by their approach in worst case \cite{Ward.2016}. To investigate the use of LiDAR for localization, Hata suggests using LiDAR to detect curbs and road markings to create a feature map of the environment and localize vehicles with the help of RTK-GPS and IMU within the map \cite{Hata.2016}. Another alternative is using ultrasonic sensors, proposed by Jung \cite{Jung.2009}. Interestingly, as the development of the internet of things (IoT), more and more researchers are focusing on SLAM by cooperative localization techniques. The basic idea of this approach is to get crucial information by adverse weather or obstacles from infrastructure or other vehicles.  For example, del Peral-Rosado showed the feasibility of 5G based localization technology \cite{delPeralRosado.20166282016630}, and Rohani utilized VANET to enhance GPS accuracy to 3.3m mean level \cite{Rohani.2015}. Further information can be found in the survey paper from Kutti \cite{Kuutti.2018}.

\subsubsection{Construction machines}

for mobile construction machines, the requirements for localization techniques can be different based on different use cases. Some machines may work in the underground, where the situation is similar to working in a tunnel, whereas others might work on an open-pit mining site. In underground mines field, it was proposed to use the laser for extracting the wall positions, and dynamically generate a path from these laser data while considering variable offsets \cite{Larsson.2010, Roberts.2002}. In contrast, for the open-pit site, an autonomous wheel loader introduced by Gu \cite{Gu.2018} uses a set of sensors, including GPS and IMU for localization, and LIDAR, radar as well as a camera for obstacles capture and identification, ensuring it perceives surroundings accurately. Moreover, Xiang created a dataset for mobile machines detection from the view of a camera fixed on the ground \cite{Xiang.2020g}, while Bang proposed a method recognizing the machines from a view of the drone \cite{Bang.2019}. In additon, the visual SLAM is proposed \cite{Shang.2019}. Besides that, V2X technology was introduced in the field of construction machines \cite{Zhou.2019, Zhou.2017}. In 2020, Xiang proposed to use WIFI to achieve the communication between different vehicles by introducing a realtime estimation method with respect to package loss and delay \cite{Xiang.2020e}. Afterward, the feasibility of using 5G for machines is also investigated by \cite{Xiang.2020h}. To avoid additional costs for the vehicles, smartphones show great potential to be utilized as a solution to complement the flaws of onboard ECU \cite{Samie.2016, Xiang.2020f}.  

In summary, similar to general autonomous vehicles, most automated mobile machines fuse information from onboard sensors such as IMU and GPS by using diverse sensor fusion technologies. Furthermore, camera, LIDAR, and radar are used to detect the environment in the construction site, to avoid obstacles, and to instruct the machines where to go. However, owing to the harsh environment and diversity of working sites, LIDAR and radar can be sensitive. In the recent future, wireless communication can also contribute to better localization of vehicles.

\section{Model building}
The wheel loader used in the simulation was modeled in Solidworks and then imported into the Robot Operating System (ROS) to explore the approaches that should be used for mobile machines.  Since the first goal of this paper is to find out  suitable sensor arrangements to accurately localize wheel loaders, we fused different arrangements of sensor data. Concretely, we use up to three IMUs and three GPSs in the simulation.
Based on the characteristic of GPSs, three GPSs are fixed on the cab of the wheel loader. We then installed two IMU sensors under the front axle and other IMUs under the rear axle. Fig. \ref{fig:wheel loader model in Gazebo} illustrates the wheel loader model we used in the Gazebo environment.

\begin{figure}[htbp]
    \centering
    \includegraphics[width=3.5in]{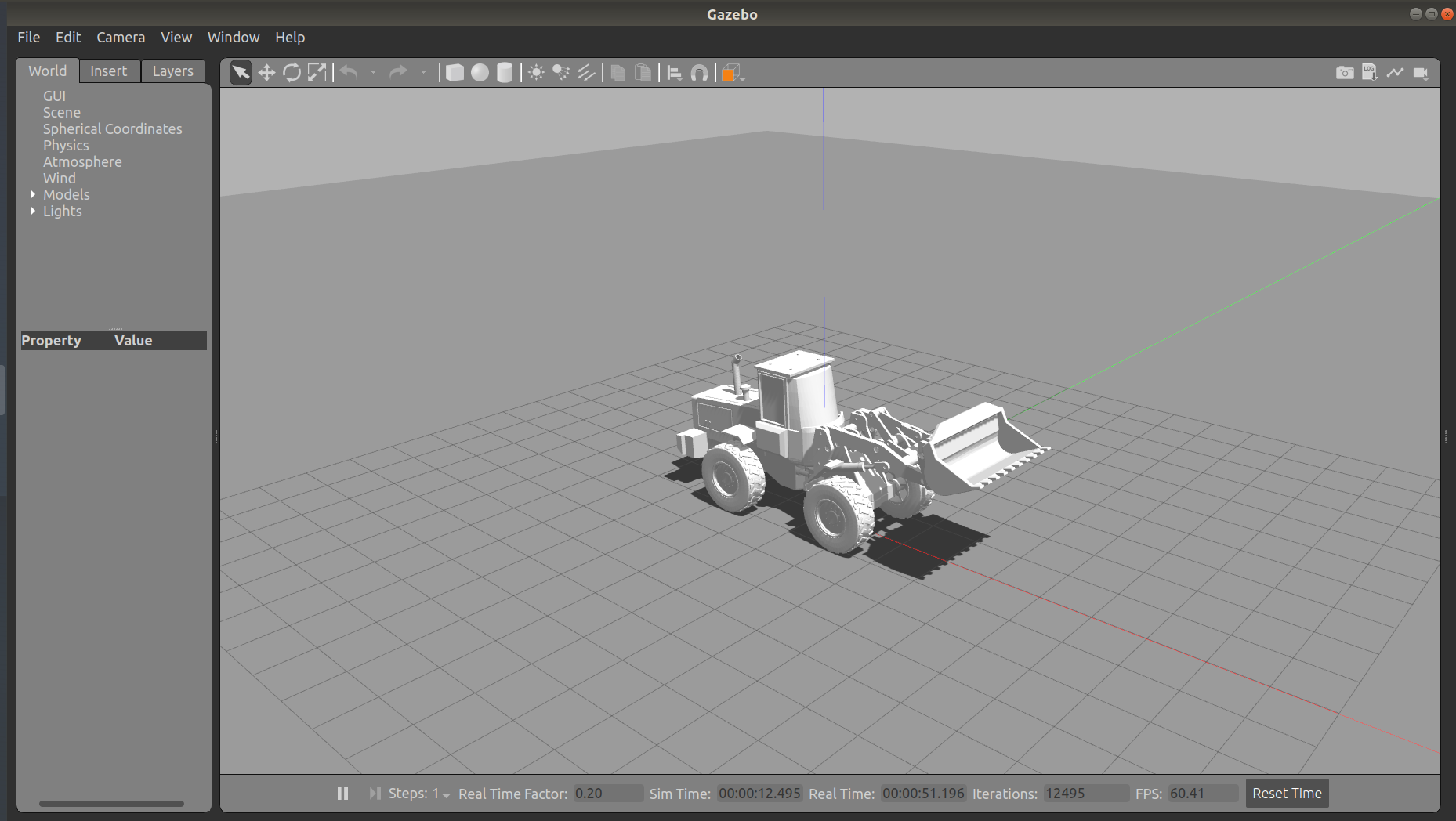}
    \caption{Wheel loader model in Gazebo: once the models had been developed in Solidwork, they were converted to Unified Robotic Description Format (URDF), using a 3rd party URDF conversion tool called ``\emph{sw\_urdf\_exporter}”, which allows for conveniently export SW Parts and Assemblies into a URDF file. Gazebo enables us to obtain sensors’ simulation such as IMUs, GPSs, encoders, cameras, and stereo cameras through \emph{gazebo\_plugins}, which can be used to attach into ROS messages and service calling the sensor outputs, i.e., the \emph{gazebo\_plugins} create a complete interface (Topic) between ROS and Gazebo.}
    \label{fig:wheel loader model in Gazebo}
\end{figure}

\begin{figure*}
    \centering
    \includegraphics[width=7in]{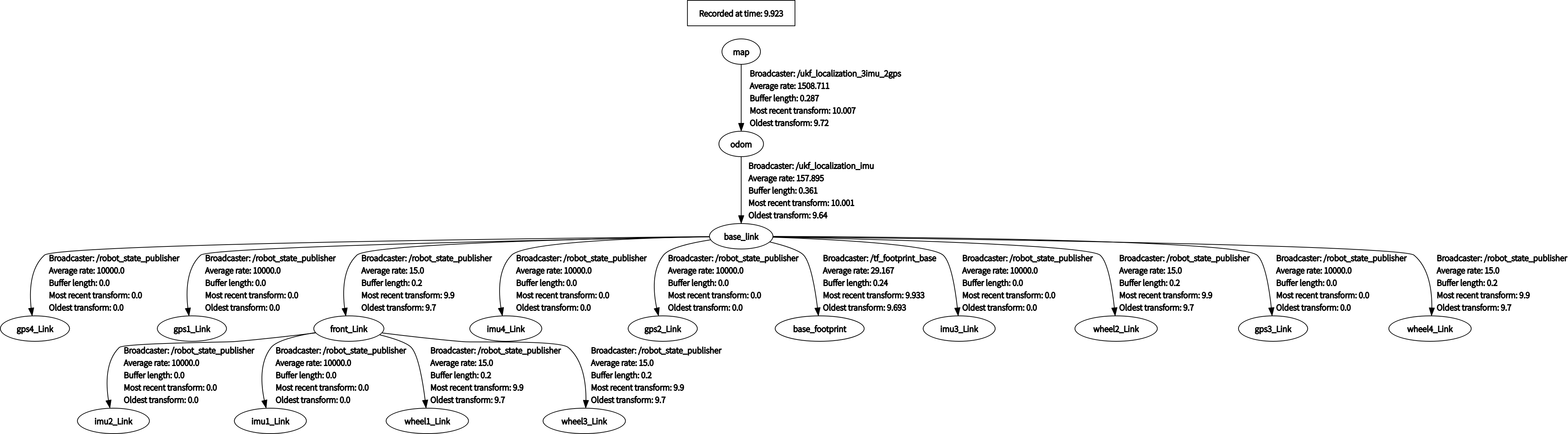}
    \caption{The dynamic system simulated by URDF file on ROS}
    \label{fig:The dynamic system simulated by URDF file on ROS}
\end{figure*}

As we know, the URDF is an XML file format used in ROS to describe all elements of a vehicle. URDF can specify the kinematic and dynamic properties of a single robot in isolation. To make our vehicle works properly in Gazebo, additional simulation-specific tags concerning the vehicle pose, frictions, inertial elements, and other properties have been added. The transform tree is shown in Fig. \ref{fig:The dynamic system simulated by URDF file on ROS}.

Each Link in URDF represents a rigid body. Also, according to the kinematic and dynamic model shown in Fig. \ref{fig:The dynamic system simulated by URDF file on ROS}, the wheel loader in the simulation is divided into several parts, including,
\begin{itemize}
    \item [1.]	\emph{base\_link} represents the rear part of the wheel loader, which is also the main coordinate frame of the simulated model in ROS, because most of the sensors are attached to the rear part and considered as child links.
    \item [2.]	\emph{front\_link} represents the front part of the wheel loader, considered as the child link of the \emph{base\_link}, and connected by a revolute joint.
    \item [3.] 	\emph{wheel\_link} represents the wheels of the wheel loader. Apparently, the two front wheels are attached to the front part, and the two rear wheels are attached to the rear part.
    \item [4.] 	\emph{gps\_link}: the GPS devices, which fixed on the roof of the wheel loader.
    \item [5.] 	\emph{imu\_link}: the IMU devices, which two of them fixed under the front part of the wheel loader and other two fixed under the rear part of the wheel loader.
\end{itemize}

In this project, our wheel loader receives GPS data from an onboard GPS sensor plugin with its latitude and longitude; however, the GPS data provided by the GPS plugin cannot be directly applied to the fusion of the sensor data, so coordinate system conversion for GPS data is required. For our simulation, we define a transform for each GPS that converts the vehicle’s world frame coordinates, i.e., the frame with its origin at the vehicle’s initial position, to the GPS’s UTM coordinates, the same as \cite{Moore.2016}, as

\begin{equation}
    \mathbf{\mathrm{T}} = \begin{bmatrix}
    c \theta c \psi & c \psi s \phi s \theta-c \phi s \psi & c \phi c \psi s \theta + s \theta s \psi & x_{{UTM}_0}\\
    c \theta s \psi & c \phi c \psi + s \phi s \theta s \psi & -c \psi s \phi + c \phi s \theta s \psi & y_{{UTM}_0}\\
    -s \theta & c \theta s \phi & c \phi s \theta & z_{{UTM}_0}\\
    0 & 0 & 0 & 1 \end{bmatrix}
\label{eq:3.19}
\end{equation}
where $\phi$, $\theta$, $\psi$ denote the vehicle’s initial UTM-frame roll, pitch, and yaw. c and s denote the cosine and sine functions, respectively. $x_{{UTM}_0 }$, $y_{{UTM}_0 }$, and $z_{{UTM}_0 }$ are the UTM coordinates of the first reported GPS position. Every time after that, we transform the GPS signal into the vehicle’s world coordinate frame, \emph{odom}, by

\begin{equation}
    \begin{bmatrix}
    x_{odom}\\
    y_{odom}\\
    z_{odom}\\
    1
    \end{bmatrix}
    =
    {\mathbf{\mathrm{T}}}^{-1}
    \begin{bmatrix}
    x_{{UTM}_t}\\
    y_{{UTM}_t}\\
    z_{{UTM}_t}\\
    1
    \end{bmatrix}
\label{eq:3.20}
\end{equation}
In our simulation the ROS package ``\emph{robot\_localization}" from Moore \cite{Moore.2016} was used, including a ``\emph{navsat\_transform}" node, which provides functions to convert between various coordinate frames and integrate GPS data. It provided a transformation function that allows the conversion between GPS frame, expressed in latitude and longitude, and vehicular coordinate. This process shall be carried out for each GPS independently.

In practical applications, GPS signals can be received infrequently. Yet the localization technology must maintain state estimation even when some of the vehicles' signals are absent. Therefore, we evaluate the performance of the filters when GPS signals infrequently arrive in our environment. Taking this problem into consideration, we build a ROS node to filter the collected GPS signals such that GPS data is unavailable for 1 second once every 10 seconds. Since we use three GPS, the signal failure might not happen at the same time. We aim to observe how the filter and our approach behave with different sensor configurations when some GPS signals go wrong.

\subsection{Sensor fusion for localization}
\label{Sensor fusion for localization}
For localization of the wheel loader in the simulation environment, we use EKF and UKF node in ``\emph{robot\_localization}"  \cite{Moore.2016}. On the one hand, this package has no limitation for the number of sensor inputs, which just in line with our construction machines' requirements. On the other hand, a concrete motion model is not needed so that this method can easily be used on both excavators and wheel loaders with different drivetrain solutions. In ``\emph{robot\_localization}", the filter’s state will be driven forward by a standard 3D kinematic model derived from Newtonian mechanics to calculate the vehicle's motion, including position, velocity, and acceleration in three dimensions.

In the correction step, the measurement model integrates sensor data to update the predicted state. GPS provides position information, and wheel encoders provide velocity information for the correction. Moreover, the orientation, velocity, and acceleration information are updated from IMUs via gyroscopes and accelerometers. Furthermore, the process model and measurement model also need to add a noise covariance matrix, Q and R, respectively. The noise matrices can generate uncertainty in the system. The process matrix contributes to the overall uncertainty in the algorithm, which adds to the process model. Intuitively, a large value in the Q matrix means a considerable uncertainty in the process model, causing the system to have greater confidence in the measurement data. In the current implementation, the process noise matrix was set as a diagonal matrix. The state variables, which are directly measured by the sensors, such as x y position by GPS and orientation by IMUs, were set relatively small. The variables, which were not directly measured, could be updated from the measured data.
The measurement covariance matrix R corresponding to the confidence in the sensor data. Similarly, the greater the noise in the elements of this matrix, the less confidence in the measurement data. The measurement covariances are derived by the sensors noisy.

In addition to the inaccuracy of filters, outliers are also an important source of error. In our simulation, we assume that the measurements have Gaussian distributions. Although sensors follow the normal distribution as setting, improbable, and extremely noisy measurements can appear due to the high fidelity of ROS. To counteract this problem, we use Mahalanobis distance to detect outliers and thus overcome the consequent adverse effect. After this, the filtered data will be used for state correction.
Concretely, the Mahalanobis distance is calculated as a product of the innovation vector to find out the outliers,
\begin{equation}
    D_M(\vec{x}) = \sqrt{{(Z_t-\hat{Z}_t)}^{\mathrm{T}}A^{-1}(Z_t-\hat{Z}_t)}
\end{equation}
where A is the covariance matrix.


\subsection{Sensor fusion methodes}
State estimation is one of the most critical issues in many autonomous applications.  Having an accurate state estimation, we can effectively navigate the machines in the environment and thereby making optimal decisions for specific purposes. For instance, to reach a target destination, it needs to know its current state, which consists of position, velocity, acceleration, and heading to execute the right maneuvers correctly. Since sensors are susceptible to noise and imperfections introducing uncertainty to the measurements, the filter's goal is to fuse all the available sensor data, as well as the vehicle’s own dynamics to obtain a more precise estimation of the vehicle’s state. As mentioned, we present two necessary extensions of the Kalman filter, notably the EKF and UKF.

The filters are modeled to improve the positioning's accuracy by compensating for the disadvantages of the different sensors. As we know, GPS provides relatively accurate positioning, but the signal’s availability remains a problem, especially in urban and mountainous environments. This determines that the results of using GPS alone are usually not satisfactory. Also, IMUs use a combination of accelerometers and gyroscopes to measure linear accelerations and angular velocities, respectively. By estimating the position relative to its initial position, the trajectory can be calculated by these information of the wheel loader as the vehicle travels. For sure, there is also a common problem with IMUs, namely the accumulated errors. To overcome this problem, we shall correct the estimated position using other sensors, to avoid accumulated drift and provide global positioning. Without a doubt, a GPS/IMU system is successful in increasing the accuracy beyond standalone GPS or IMU capabilities.

\subsubsection{Extended Kalman Filter}
although linear Gaussian systems are abundant, most systems, in reality, are non-linear. Also, they often do have Gaussian noise. Wrong assumptions about the system can lead the Kalman filter to diverge and provide estimation with very high errors. Consequently, multiple extensions have been developed to deal with various scenarios encountered in practice. One of the famous variations is the EKF, where it deals with non-linearity by approximating a linear equivalent before performing the required filtering sequence. The idea of the EKF is that if the system is close to linear for short periods, using its linear approximation will then not yield large errors.

Through linearizing the basic equations form Welch \cite{Welch.1995}, we get the following equations:

\begin{equation}
x_{k+1}\approx{\tilde{x}_{k+1}}+A(x_k-{\hat{x}}_k)+W{w_k}\label{eq:2.9}  
\end{equation}
\begin{equation}
z_k\approx\tilde{z}_k+H(x_k-{\hat{x}}_k)+Vv_k  \label{eq:2.10}  
\end{equation}
where $x_{k+1}$ and $z_k$ are the actual state and measurement vectors, $\tilde{x}_{k+1} $ and $\tilde{z}_k $ are the approximate state and measurement vectors. $ {\hat{x}}_k $ is an \emph{a posteriori} estimate of the state at step $k$, random variables $w_k$ and $v_k$ represent the process and measurement noise. $A$ is the Jacobian matrix of partial derivatives of $f(\bullet)$ with respect to $x$, $W$ is the Jacobian matrix of partial derivatives of $f (\bullet) $ with respect to $w$, $H$ is the Jacobian matrix of partial derivatives of $h (\bullet)$ with respect to $x$, $V$ is the Jacobian matrix of partial derivatives of $h (\bullet)$ with respect to $v$.

An essential feature of the EKF is that the Jacobian $H_k$ in the equation for the Kalman gain $K_k$ serves to correctly propagate only the relevant component of the measurement information \cite{Welch.1995}; the linearization error is always exist as the function is nonlinear. Because increasing the sampling time and reducing the nonlinearity of function are not always viable, error-state EKF is proposed to counteract the adverse effect. The basic idea of error-state EKF is to reduce the distance of the linear approximation from the operating point; instead of linearization of the nominal state, it handles the error state.

\subsubsection{Unscented Kalman filter}\label{AA}

when the state transition and observation models, that is, the predict and update functions $f$ and $h$ are highly nonlinear, the EKF can give particularly poor performance. This is because the covariance is propagated through the linearization of the underlying nonlinear model. By contrast, the UKF uses unscented transform instead of linearization in the prediction and correction steps to make the estimation.  As the first step, we shall compute the decomposition of the covariance matrix and carefully select the sample points, for instance, here we selected 2L+1 points, described as, 
\begin{align}
{\mathcal{X}}^0 \quad  &= \bar{x} \notag  \\
{\mathcal{X}}^i \quad &= \bar{x}+(\sqrt{(L+{\lambda})P_x})_i \qquad  & i & = 1,\cdots,L\notag\\
{\mathcal{X}}^{i+L} \quad &= \bar{x}-(\sqrt{(L+{\lambda})P_x})_{i} \qquad  & i & = 1,\cdots,L\notag\\
\end{align}
where $\bar{x}$ is the selected mean sample point and $\lambda$ is set as $\lambda =3-N$ for Gaussian probability density function.  After that, the sigma points will be propagated through the nonlinear function as,
\begin{equation}
\check{X}^i = f({\mathcal{X}}^i) \qquad i = 1,\cdots,2L \label{eq:2.16}  
\end{equation}
where $f(\bullet)$ is the motion model function and and $\check{X}^i$ is the predicted position based on the motion model. As the final step of predicted process, the predicted mean and covariance should be calculated. 

\begin{equation}
a^{i} =
\begin{cases} 
\lambda/L+\lambda, i=0 \\
\lambda/2(L+\lambda), otherwise 
\end{cases}
\label{eq:mean and covariance}
\end{equation}

\begin{equation}
\check{X} = \sum\limits_{i=0}^{2N}  a^{i} \check{X}^i
\end{equation}

\begin{equation}
\check{P} = \sum\limits_{i=0}^{2N}  a^{i} (\check{X}^i - \check{X})(\check{X}^i - \check{X})^T + Q
\end{equation}
Here $Q$ denote the noisy of this process. In the correction step, we firstly calculate the predict measurement with the sigma points, 

\begin{equation}
\hat{y}^{(i)} = h( \check{X}^i,0),  i=0,...,2N
\end{equation}
where $h(\bullet)$ is the predicted measurement model, considering the process noise: $\check{L}\check{L}^T = \check{P}$. 
Then, we get the mean and covariance of predicted measurements,

\begin{equation}
\mathcal{Y} = \sum\limits_{i=0}^{2N}  a^{i} \mathcal{Y}^i
\end{equation}

\begin{equation}
\check{P_y} = \sum\limits_{i=0}^{2N}  a^{i} (\mathcal{Y}^i - \mathcal{Y})(\mathcal{Y}^i - \mathcal{Y})^T + R
\end{equation}
where $R$ is the noisy. Based on the previous result, we can compute the cross-covariance and Kalman gain, and then get the corrected  covariance and mean. 

\begin{equation}
P_{xy} = \sum\limits_{i=0}^{2N}  a^{i} (\check{X}^i - \check{X})(\mathcal{Y}^i - \mathcal{Y})^T
\end{equation}

\begin{equation}
K = P_{xy} P_y^{-1}
\end{equation}

\begin{equation}
\hat P = \check{P} - K P_y K^T
\end{equation}

\begin{equation}
\mathcal{X}^i = \check{X} - K(y^0 - \mathcal{Y}^i)
\end{equation}
where $y^0$ is the measurement result with respect to $\check{X}$, $\mathcal{X}^i$ is the final results from UKF. As we can see, UKF does not need Jocobian matrix so that it can achieve a better performance in estimation. Usually, the computation effort can be slightly higher than the other methods, making carefully select sensors configurations meaningful.

\subsection{Realtime map plotter}
\label{Real-time plotter}

Inspired by the research from Fankhauser \cite{fankhauser.2016}, a map can include many layers to store different types of data information. Thus, to develop a realtime plotter of the construction site according to ground condition, we use a multi-layer grid-based map, which divides the environment into uniform cells. 
Fig. \ref{fig:A grid-based map uses multilayered grid maps to store data for different types of information} illustrates the multilayered grid map concept, where each cell data is stored on the congruent layers. In this project, since resistance and grade of the road are the most of importance information for the construction machines, we adopt a two layers grid map. Concretely in our study, the map is divided into small cells, whose resolution is 1 meter per cell. As we discussed in previous section, we use GPS/IMU fusion Kalman filter algorithms to locate the mobile machine in the construction site. 

As can be seen in Fig. \ref{fig:A grid-based map uses multilayered grid maps to store data for different types of information}, to describe the ground condition of construction sites, we use the values of each cell to represent the information of the ground situation. In Fig. \ref{fig:One-layer grid-based map}, a layer that holds the data of a grid-based map is shown. Obviously, although we only demonstrate the approach with two layers,  it is relatively easy to extend the third layer in case that uphill or downhill is vital, as the third layer is responsible for recording the heading of the vehicle.  

\begin{figure}
    \centering
    \includegraphics[width=3.5in]{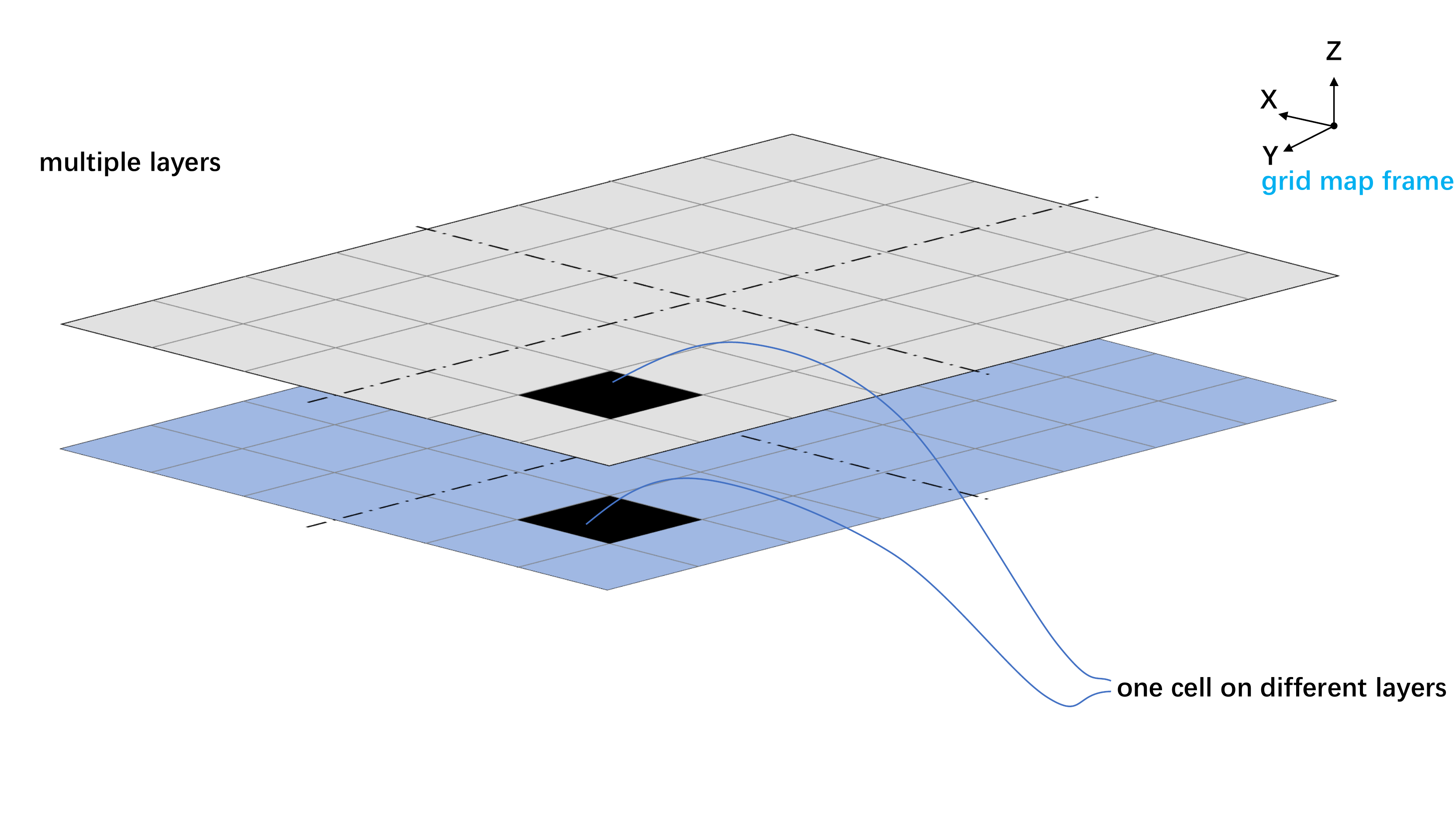}
    \caption{Our approach uses multilayered grid maps to store data for different types of information. Concretely, every gird saves a 1*3 matrix including location information and resistance or grade, depending on which layer it is. A gird with site information will be created after the vehicle passes by. The map is saved as a  2*m*n*3 tensor, where m is the max displacement in the x-direction, while n is the max displacement in the y-direction. In case a gird does not be occupied once by the vehicle, it will be marked as NaN to denote the unknown regions.}
    \label{fig:A grid-based map uses multilayered grid maps to store data for different types of information}
\end{figure}

\begin{figure}
    \centering
    \includegraphics[width=3.5in]{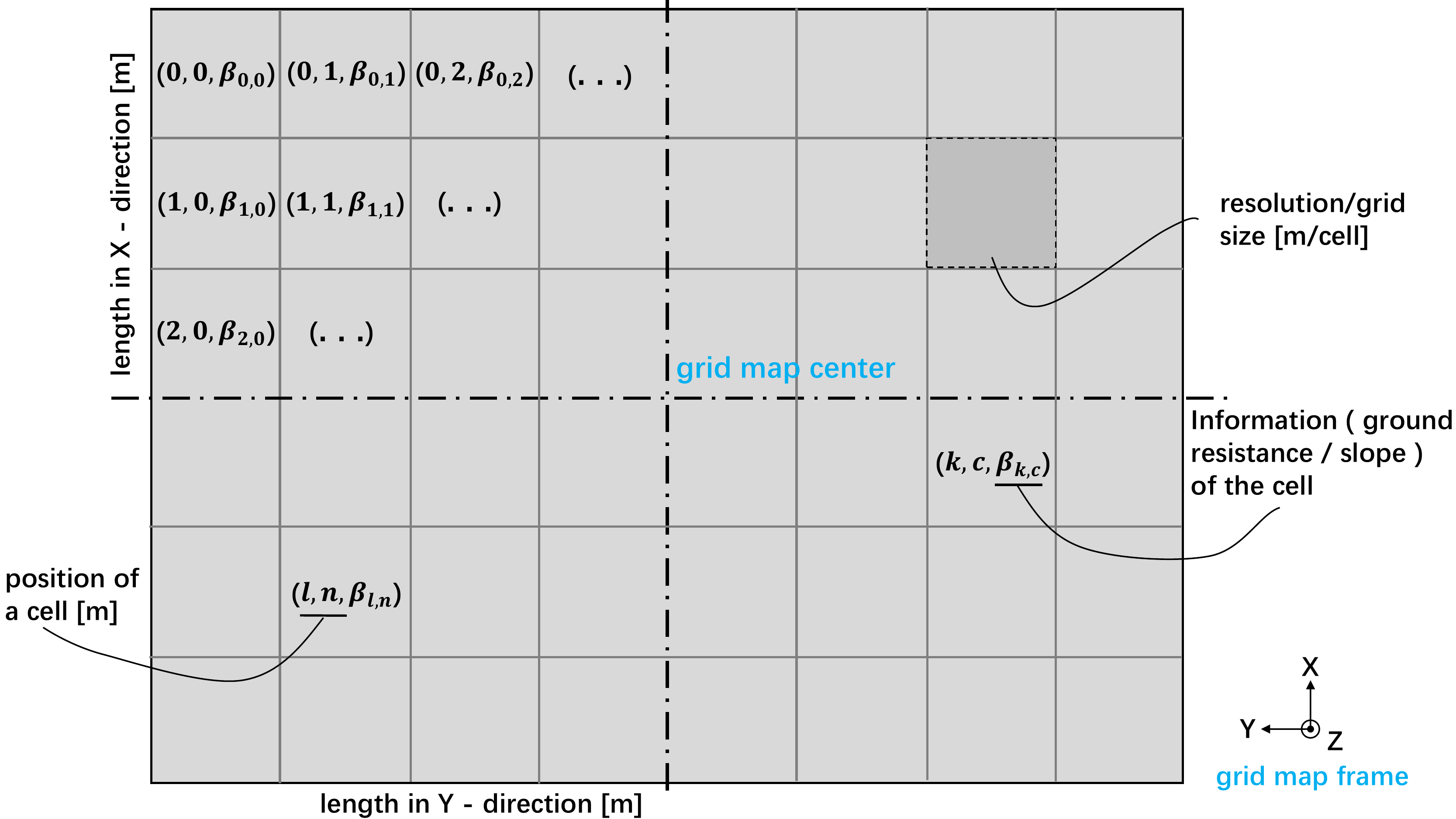}
    \caption{Detail description of a layer in the grid-based map}
    \label{fig:One-layer grid-based map}
\end{figure}

Based on the previous study, both resistance and grade of the road can be gathered in realtime. Thus, we assume that the ground resistance and slope are known after the mobile machine passed by. When the mobile machine passes through each grid cell,  ground information will be added to the corresponding grid of the two layer-grid-maps. Concretely, the plotting algorithm combines the localization results from the Kalman filter and the grade and resistance information from the recursive least square algorithm.  To implement the plotting algorithm in ROS, we created a \emph{node} in ROS that can subscribe to the localization results from the Kalman filter node and gather the current information from ground truth maps with the assumption that resistance and grade of the road can be estimated or measured well.

\section{Vehicle simulation scenarios in ROS and Gazebo}

To test the feasibility of the map plotter, we first define different road conditions in a construction site. ROS is a mature and flexible framework for robotics programming, providing the required tools to easily access sensors data, process that data, and generate appropriate responses for the robot's actuators. Due to these characteristics, ROS is a perfect tool for many types of research on modern robotics. After all, a mobile machine can be considered as just another type of robotics, so the same types of programs can be used to develop advanced construction machines. In this paper, we simulated a construction site showed in Fig. \ref{fig:Example construction site}. According to this real construction site, we define five different ground resistances in the simulation environment, according to different ground conditions. Moreover, two areas according to different slopes. Concretely, we distinguish the different regions based on road material.   
\begin{itemize}
    \item [\textbf{1}.] \textbf{Gravel surface}: Gravel is a loose aggregation of small, variously sized fragments of rock. It has a wide range of applications in the construction industry. Therefore, gravel surface road is very common in the construction site. The rolling resistance coefficient of gravel surface is considered as 0.02.
    \item [\textbf{2}.] \textbf{Sand surface}: Sand is a type of naturally  material that is of a loose, granular, fragmented composition, consisting of particulate things such as rock, coral, shells, and so on. The rolling resistance coefficient between the sand surface and mobile machine tires is 0.250.
    \item [\textbf{3}.] \textbf{Dry dirt road}: A dirt road is a type of unpaved road made from the native material of the land surface, which is also very normal in the construction site. The typical rolling resistance coefficient of the dry dirt road is 0.040.
    \item [\textbf{4}.] \textbf{Wet dirt road}: Same as a dry dirt road, the wet dirt road is also a typical road type in the working area. The typical rolling resistance coefficient of the wet dirt road is 0.060.
    \item [\textbf{5}.] \textbf{Dry concrete surface}: Dry concrete is a normal building material and the typical rolling resistance coefficient of the dry concrete road is 0.008.
\end{itemize}

Besides, we also define two different ground slopes appropriately. Since we only use a two-layer grid map in this research, we do not discriminate uphill or downhill. 

\begin{itemize}
     \item [\textbf{1}.] \textbf{Flat area}: The slope of the ground is near 0°. In the flat area, we can let the mobile machines move faster to increase working efficiency or reduce the reserved dynamics to let the components work in an more economic region, with only a little concern of safety.
    \item [\textbf{2}.] \textbf{15° slope area}: The slope of this area is near 15°. In this area, in contrast, the mobile machines should pay more attention to the safety. 
\end{itemize}

\begin{figure}[!t]
        \newcommand{\w}{0.45}
        \centering 
        \subfloat[Example construction site divided into five areas according to different ground resistances]{\includegraphics[width=\w\textwidth]{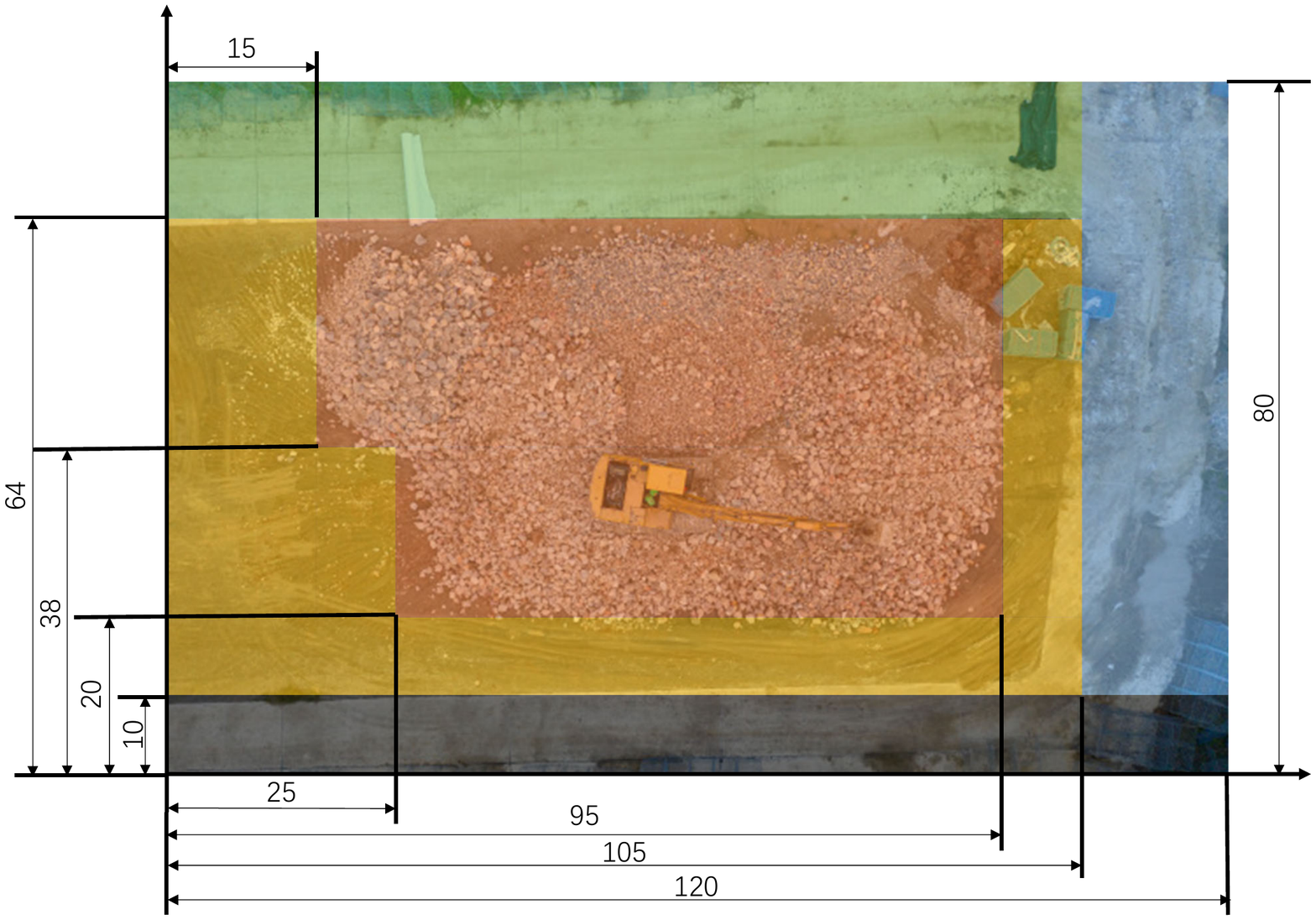}}
            \hfil 
        \subfloat[Example construction site divided into two areas according to different slopes]{
            \includegraphics[width=\w\textwidth]{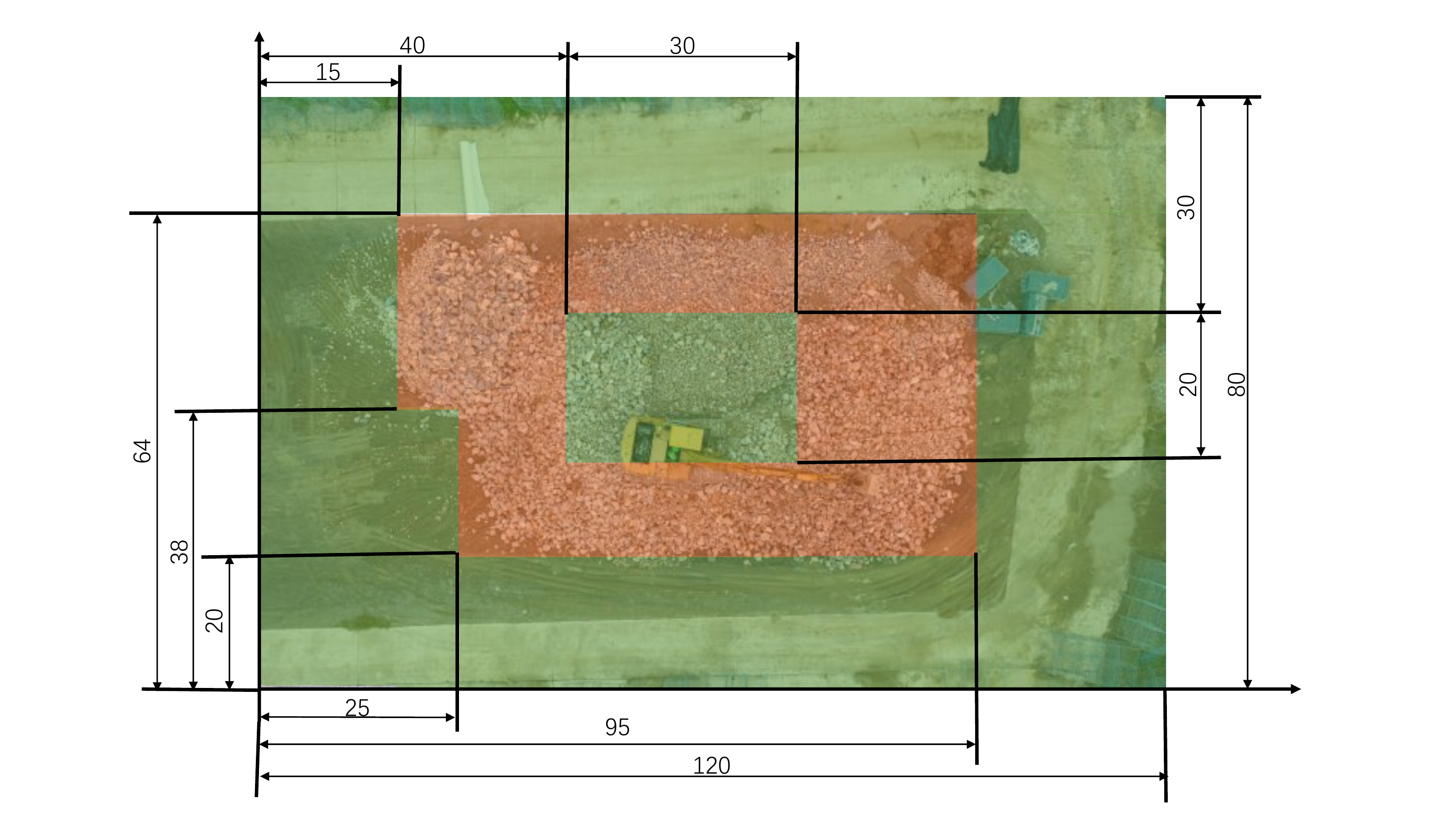}} 
            \hfil
        \caption{The ground truth map with dimensions. The simulation environment we used in Gazebo was modeled based on a real construction site, and the parameters are selected according to material characteristics. Since simulating a small construction site may cause system error and thus lack plausibility, we augmented this real construction site's dimensions in Gazebo.} 
        \label{fig:Example construction site}
\end{figure} 

For simulation, we defined the green area as the dry concrete surface in the ground-truth resistance map. The red area represents the gravel surface. The blue area represents the sand surface, the black area, and the yellow area represent dry and wet dirt roads, respectively. With this premise, we draw the ground truth map upon this construction site. Same as in the ground-truth resistances map, we also define the ground slope map, where the green area represents the flat area and the red area represents the 15° slope.

In this project, we wrote the \emph{plot\_node} with Python and OpenCV library to visualize the plotted map.  To test the feasibility of a realtime plotter in simulation, we use a \emph{ground\_truth} node in ROS, which provides a ground truth position of the simulated mobile machine in Gazebo. When the mobile machine moves to a certain position, we use the ground truth position to determine the rolling resistance coefficient and the road grade, then use the Kalman filter filtered position to plot the corresponding information in the grid-based map. 

In ROS and Gazebo, the build-in plugins provide many adjustable parameters that can be used to adjust the devices' performance. To get closer to reality, we set the performance parameters of GPSs and IMUs in the simulation according to real GPS and IMU devices.  To get the best sensor configuration, we implement different sensor configurations for different algorithms. Each group of sensor configurations is simulated under the same condition by rosbag, and the results are output simultaneously.

\section{Experiment AND RESULTS}

\subsection{Localization results}

To explore the most suitable sensor arrangements of the Kalman filter for construction machinery, we compared the results from different sensor configurations with different methods. The concrete sensor arrangements in this project are shown in Tab. \ref{tab:SENSOR CONFIGURATIONS}.

\begin{table}
    \caption{SENSOR CONFIGURATIONS}
    \scalebox{0.8}{
    \begin{tabular}{|c|c|c|c|c|c|c|c|c|}

       \hline 
       \multirow{2}*{\emph{\textbf{Group}}}&  \multicolumn{7}{|c|}{\emph{\textbf{Sensor}} ( \emph{\textbf{0}} = \emph{\textbf{deactivated}}, \emph{\textbf{1}} = \emph{\textbf{active}} )
} & \multirow{2}*{KF}\\
       \cline{2-8} ~ &  GPS 1  & GPS 2  & GPS 3  & IMU 1  & IMU 2  & IMU 3  &  Encoder & ~\\
       \hline 1 &  0  & 0 &	0 &	1 &	0 &	0 &	1 & EKF\\
       \hline 2 &  1  & 0 & 0 & 1 & 0 & 0 & 1 & EKF\\
       \hline 3 &  1  & 0 & 0 & 1 & 1 & 0 & 1 & EKF\\
       \hline 4 &  1  & 0 & 0 & 1 & 1 & 1 & 1 & EKF\\
       \hline 5 &  1  & 1 & 0 & 1 & 0 & 0 & 1 & EKF\\
       \hline 6 &  1  & 1 & 0 & 1 & 1 & 0 & 1 & EKF\\
       \hline 7 &  1  & 1 & 0 & 1 & 1 & 1 & 1 & EKF\\
       \hline 8 &  1  & 1 &	1 &	1 &	1 &	1 &	1 & EKF\\
       
       \hline 9 &  0  & 0 &	0 &	1 &	0 &	0 &	1 & UKF\\
       \hline 10 &  1  & 0 & 0 & 1 & 0 & 0 & 1 & UKF\\
       \hline 11 &  1  & 0 & 0 & 1 & 1 & 0 & 1 & UKF\\
       \hline 12 &  1  & 0 & 0 & 1 & 1 & 1 & 1 & UKF\\
       \hline 13 &  1  & 1 & 0 & 1 & 0 & 0 & 1 & UKF\\
       \hline 14 &  1  & 1 & 0 & 1 & 1 & 0 & 1 & UKF\\
       \hline 15 &  1  & 1 & 0 & 1 & 1 & 1 & 1 & UKF\\
       \hline 16 &  1  & 1 & 1 & 1 & 1 & 1 & 1 & UKF\\
       \hline 
    \end{tabular}}

    \label{tab:SENSOR CONFIGURATIONS}
    \vspace{0.1cm}
    
    In our simulation, we ignore the mirror difference caused by the slightly different installation position of sensors. Therefore, the various configurations are reduced from 64 to 16. Since odometry is robust and necessary for many applications, we do not consider the case without an encoder.
\end{table}

To evaluate the performance of the different variants of sensors and algorithms concerning accurately positioning, we controlled the wheel loader to drive on the previously defined construction site in Gazebo, and recorded the data from sensors at the same time. Afterward, the localization results of the different methods are compared to the ground truth. Here we use the root-mean-square error (RMSE) as a quantitative indicator of the error to assess the pose estimation results. 

\begin{equation}
    RMSE = \sqrt{\frac{\sum_{i=1}^n{(\tilde{S}_i- \hat{\tilde{S}}_i)}^2}{n}}
\end{equation}
Where $\hat{\tilde{S}}$ is the vector including estimated x and y position, $\tilde{S}$ is the denotes ground truth x and y position, n is number of all estimated samples, and the footnote i denotes the $i^{th} $ time step.

\begin{figure*}[!t]
        \newcommand{\w}{0.22}
        \centering 
        \subfloat[EKF with only one IMU]{\includegraphics[width=\w\textwidth]{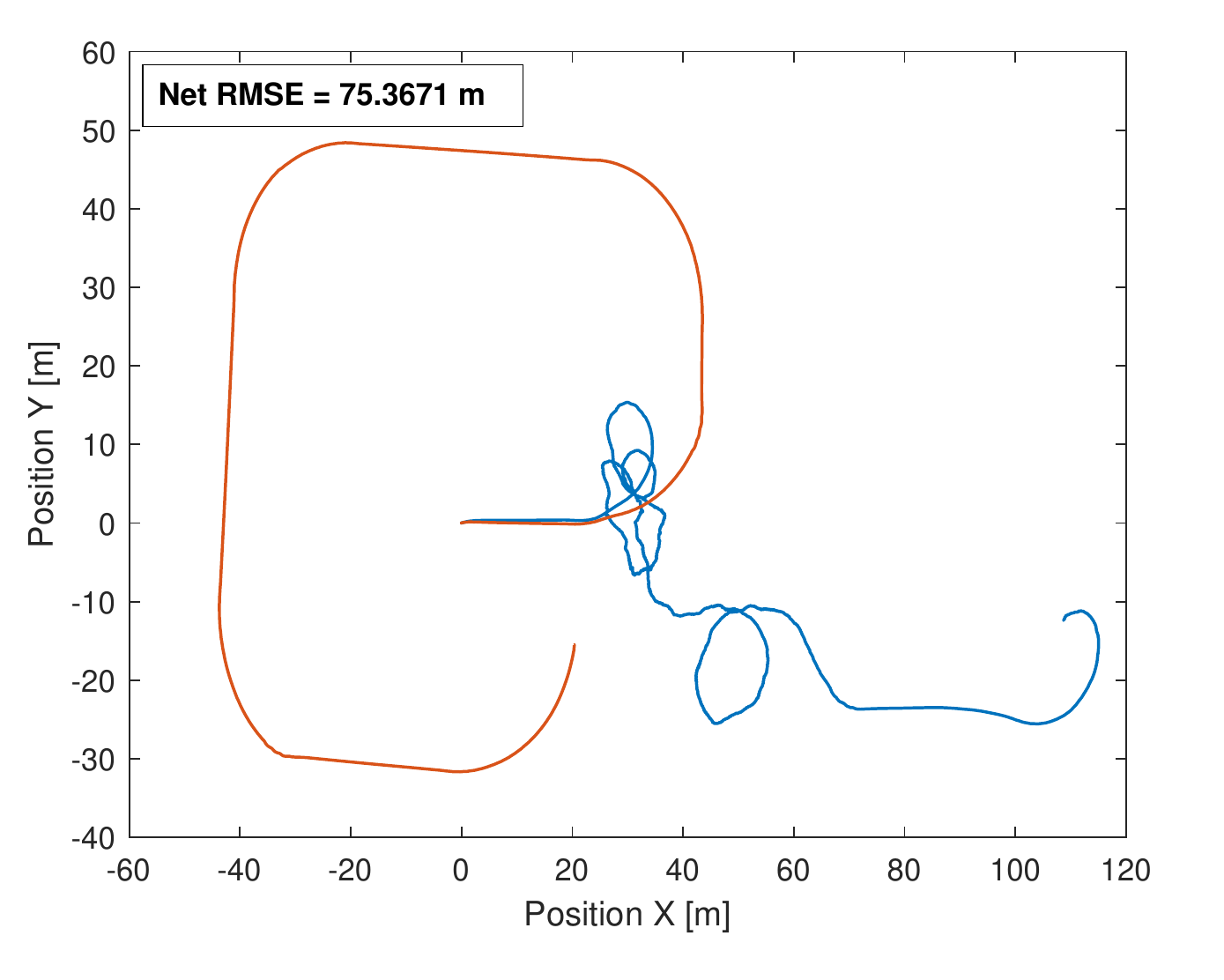}}
            \hfil 
        \subfloat[UKF with only one IMU]{
            \includegraphics[width=\w\textwidth]{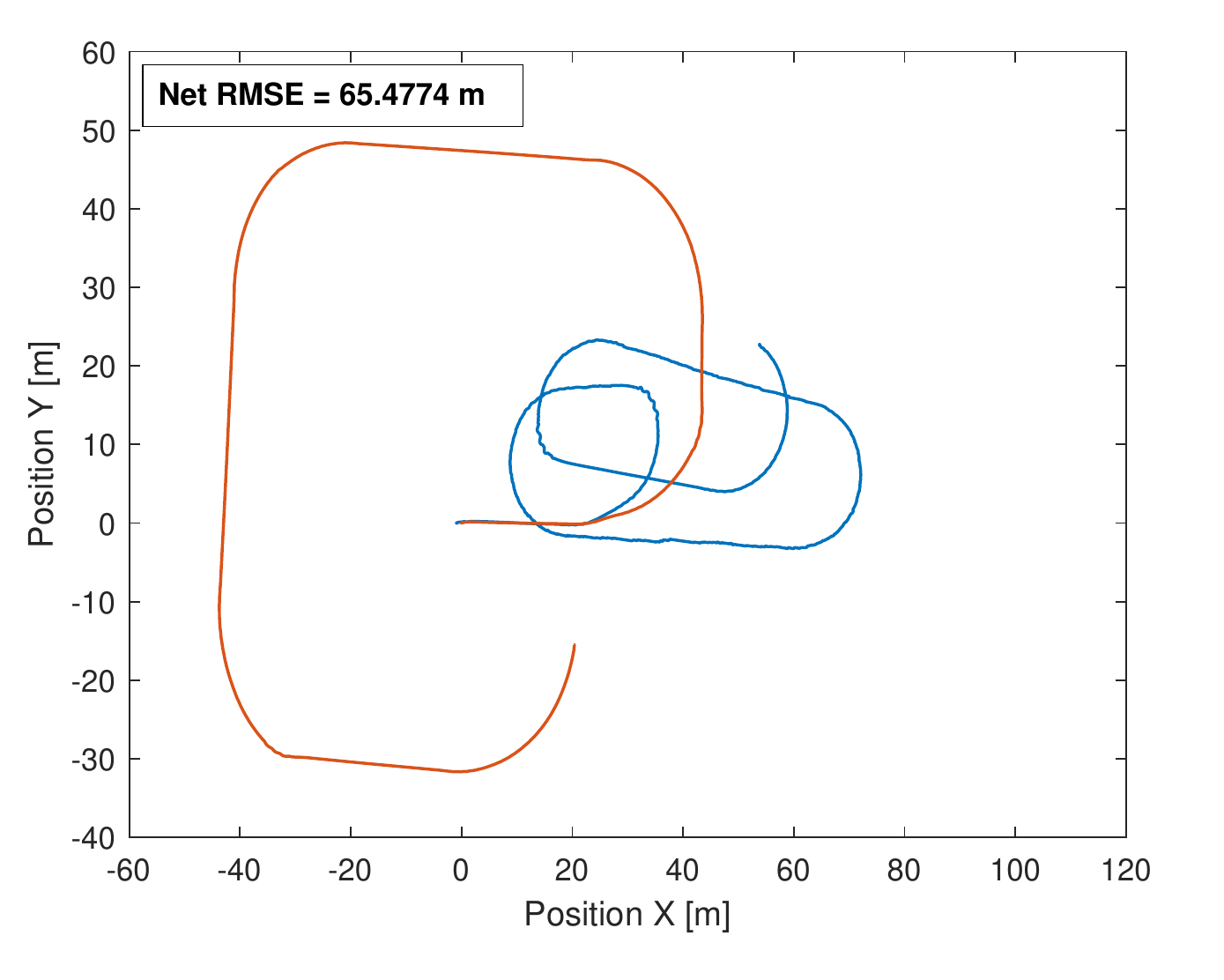}} 
            \hfil
        \subfloat[EKF with 1 IMU and 1 GPS]{
            \includegraphics[width=\w\textwidth]{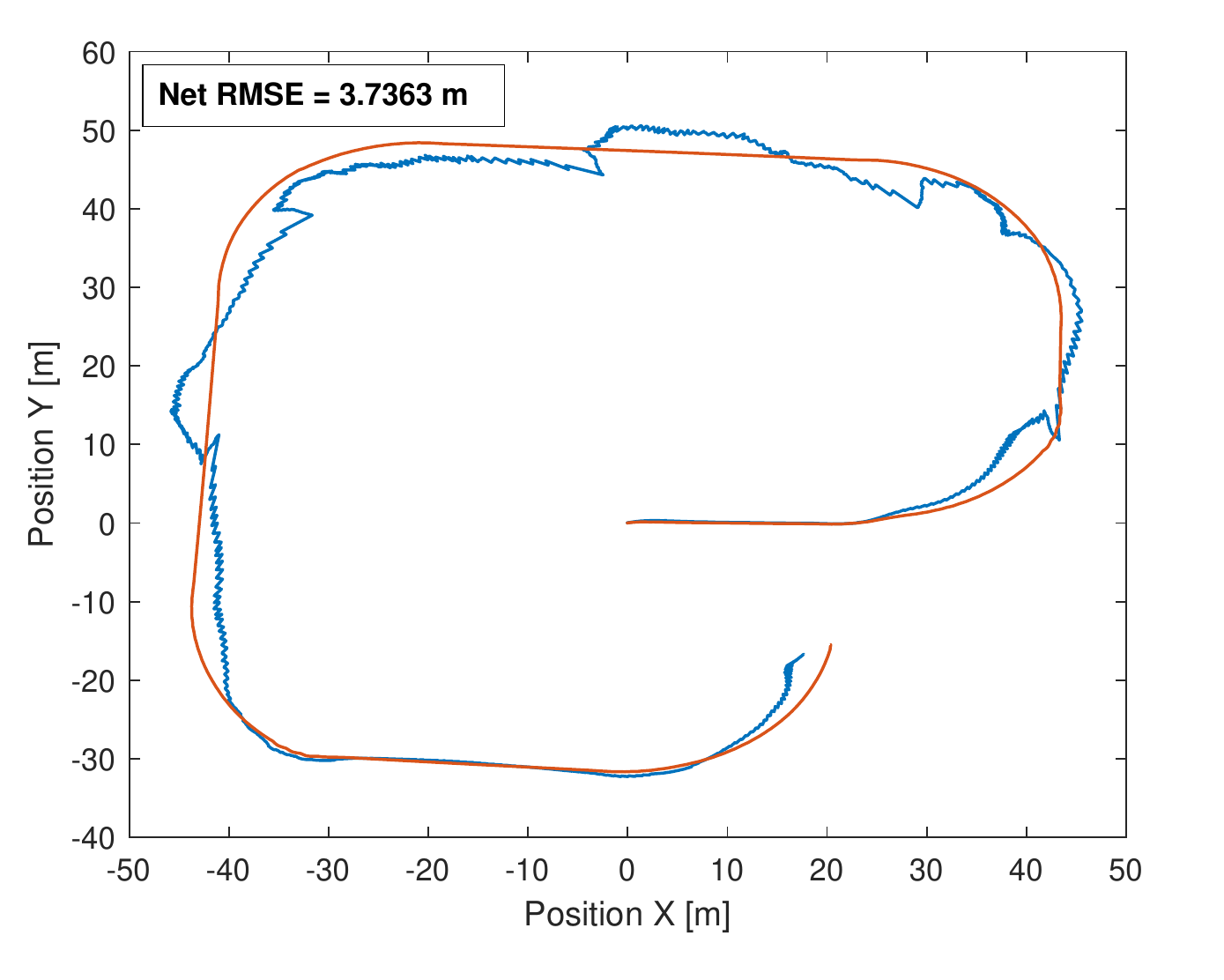}} 
            \hfil
        \subfloat[UKF with 1 IMU and 1 GPS]{
            \includegraphics[width=\w\textwidth]{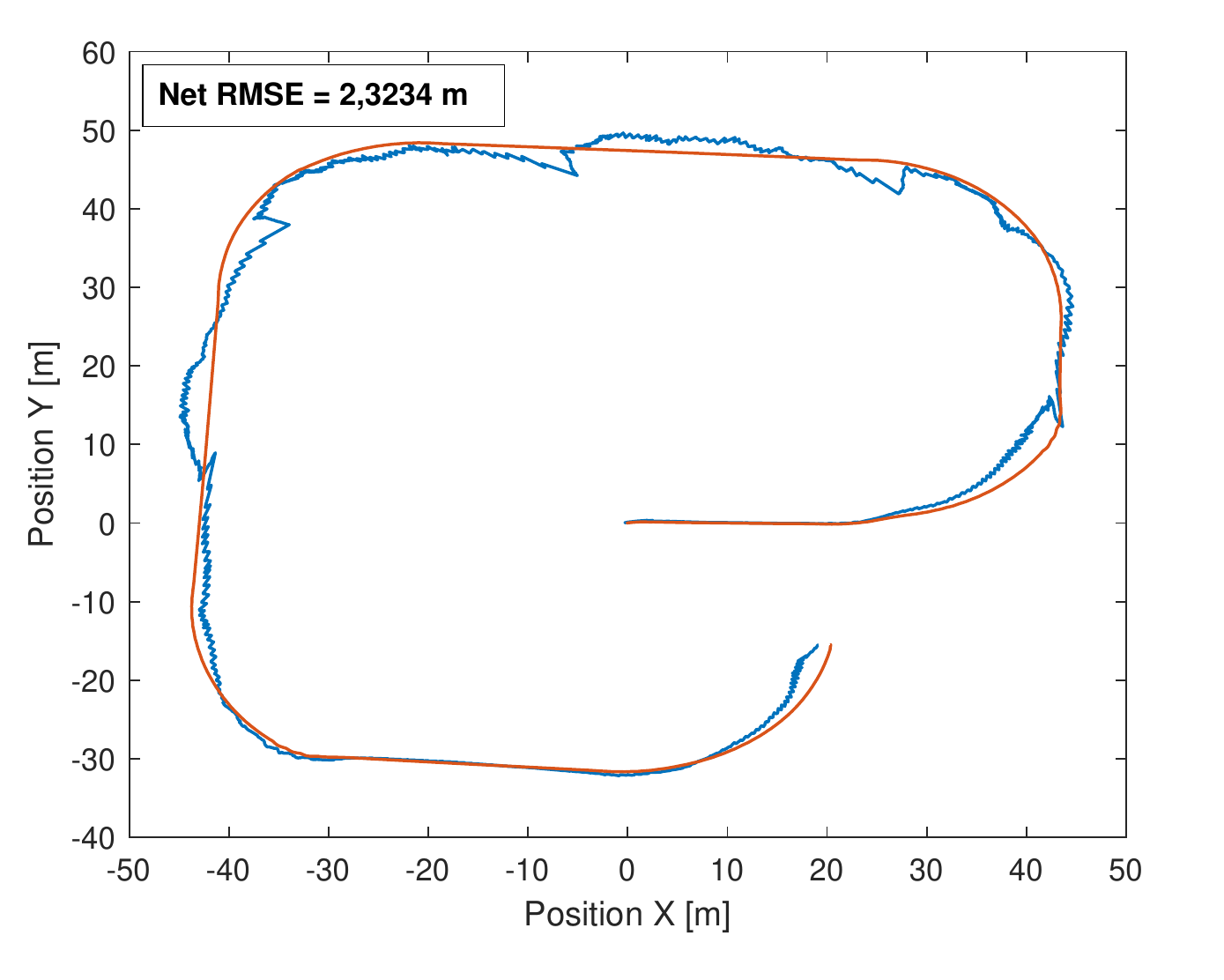}}
            \hfil 
        \subfloat[EKF with 2 IMU and 1 GPS]{
            \includegraphics[width=\w\textwidth]{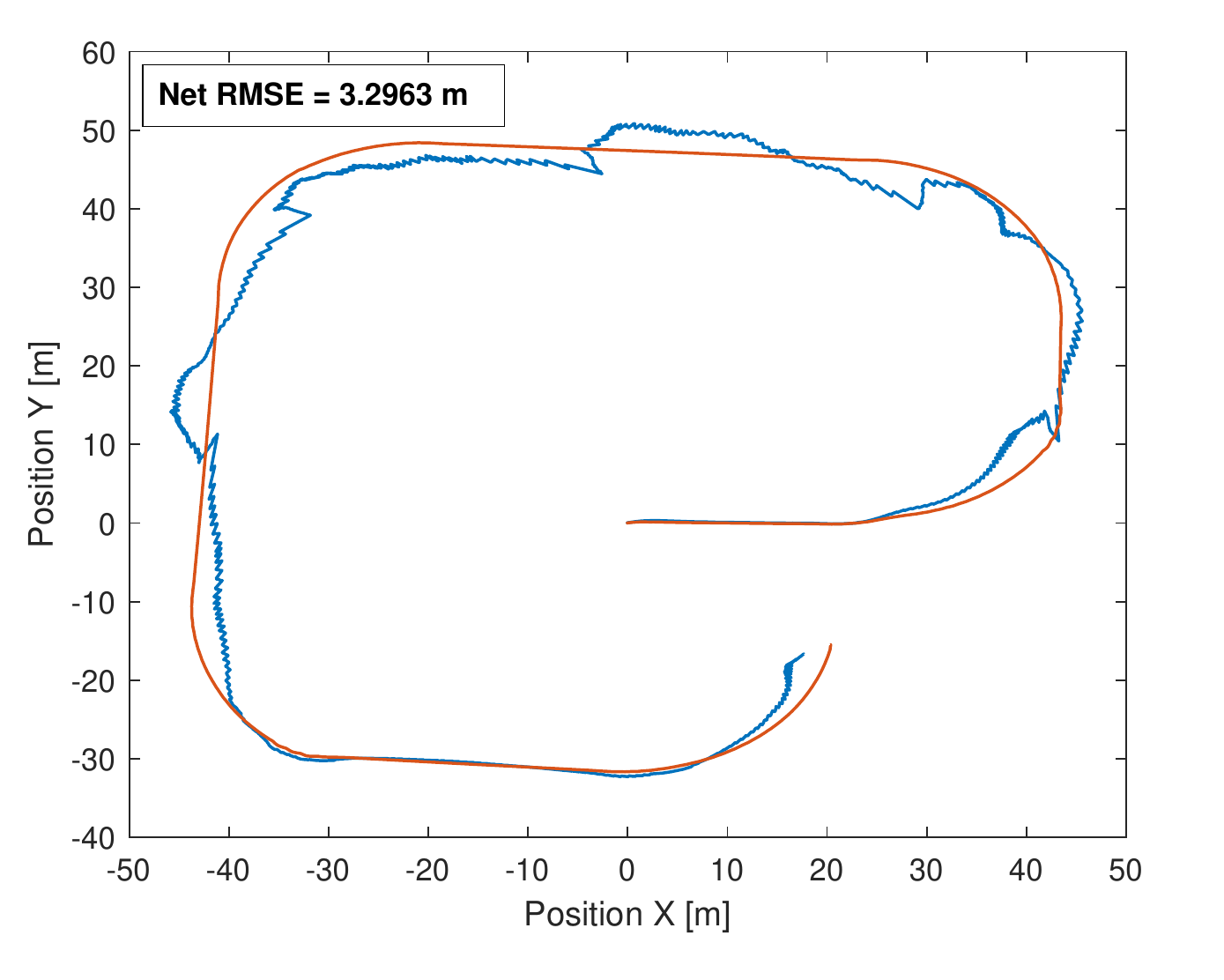}} 
            \hfil
        \subfloat[UKF with 2 IMU and 1 GPS]{
            \includegraphics[width=\w\textwidth]{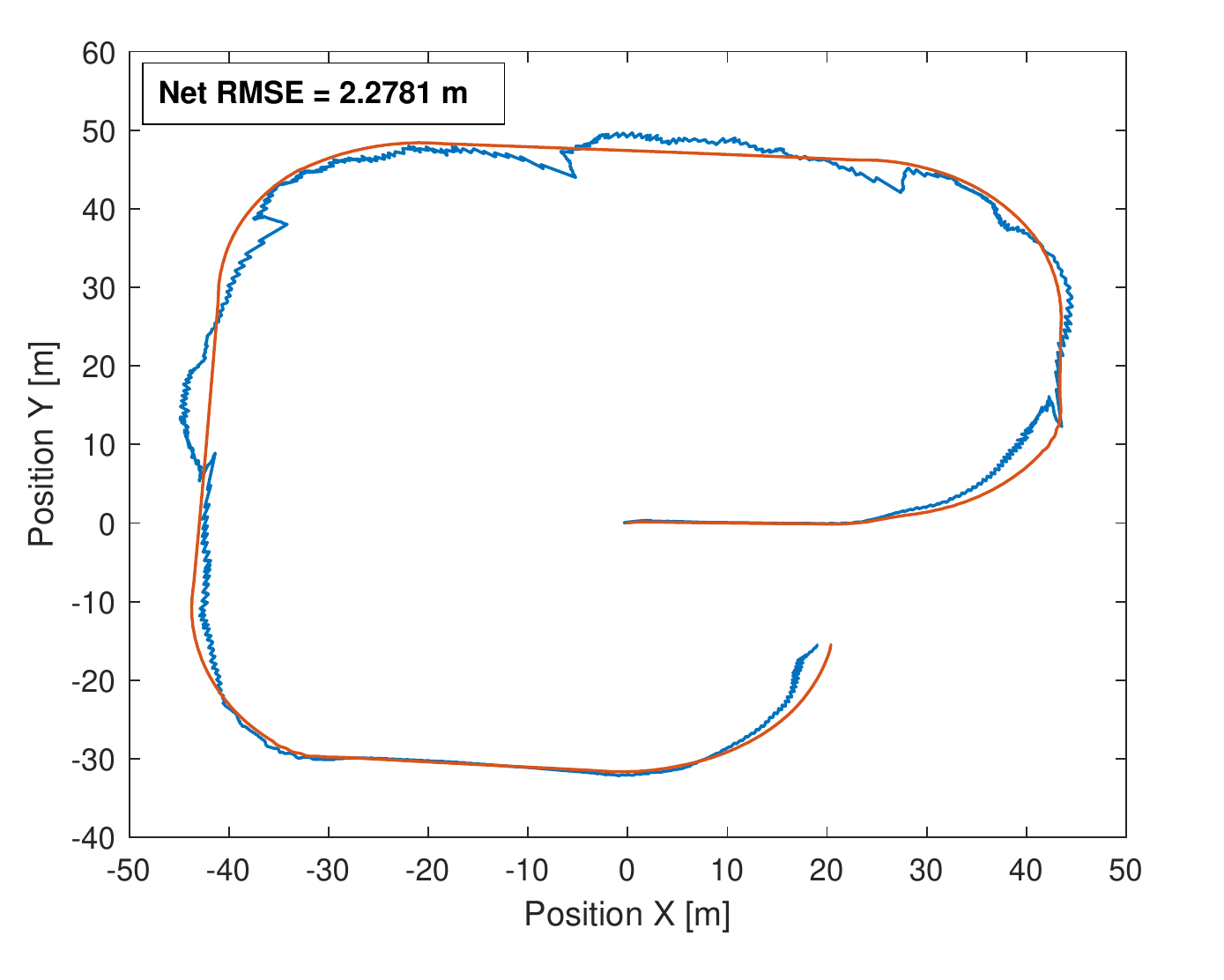}} 
            \hfil
        \subfloat[EKF with 3 IMU and 1 GPS]{
            \includegraphics[width=\w\textwidth]{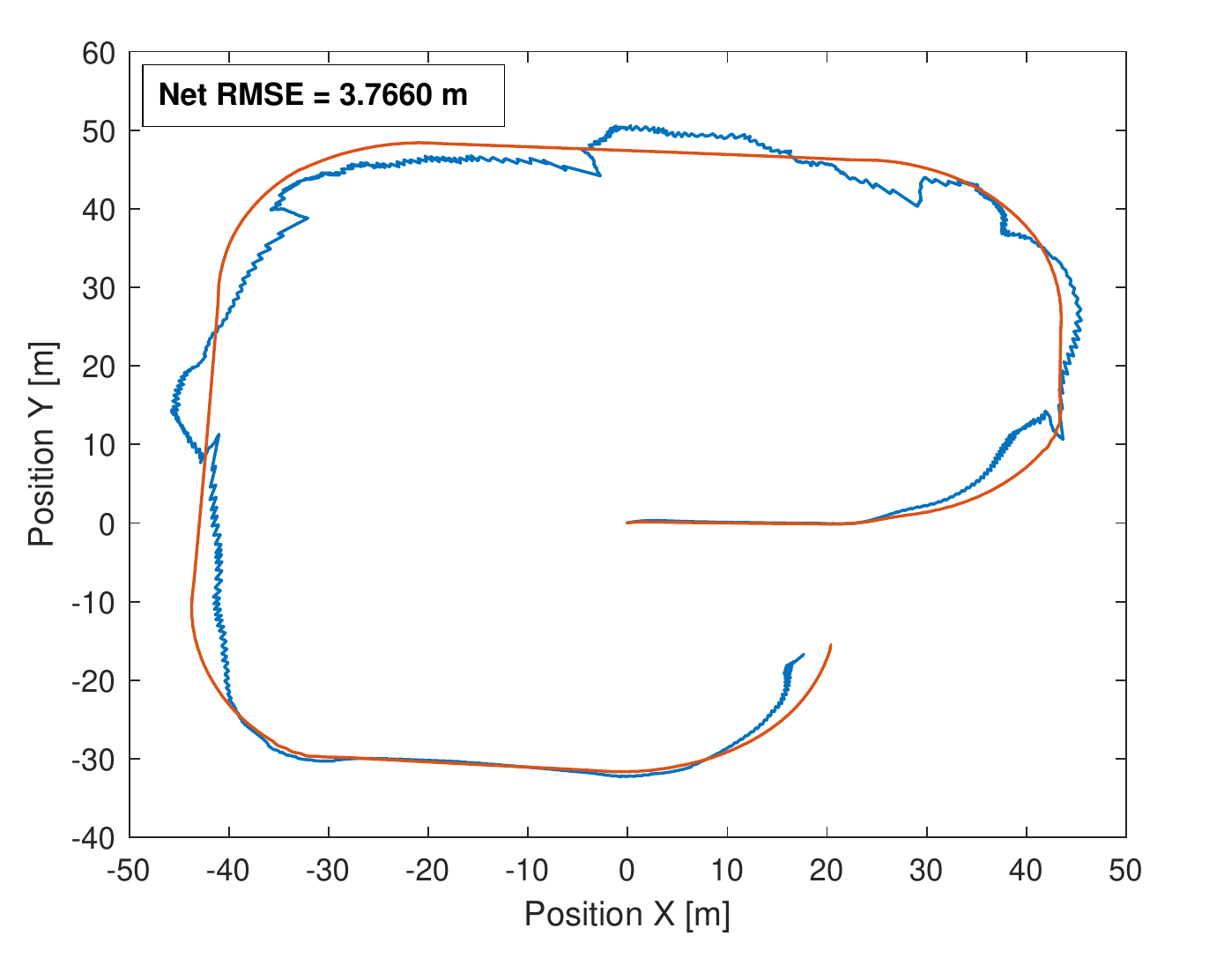}}
            \hfil 
        \subfloat[UKF with 3 IMU and 1 GPS]{
            \includegraphics[width=\w\textwidth]{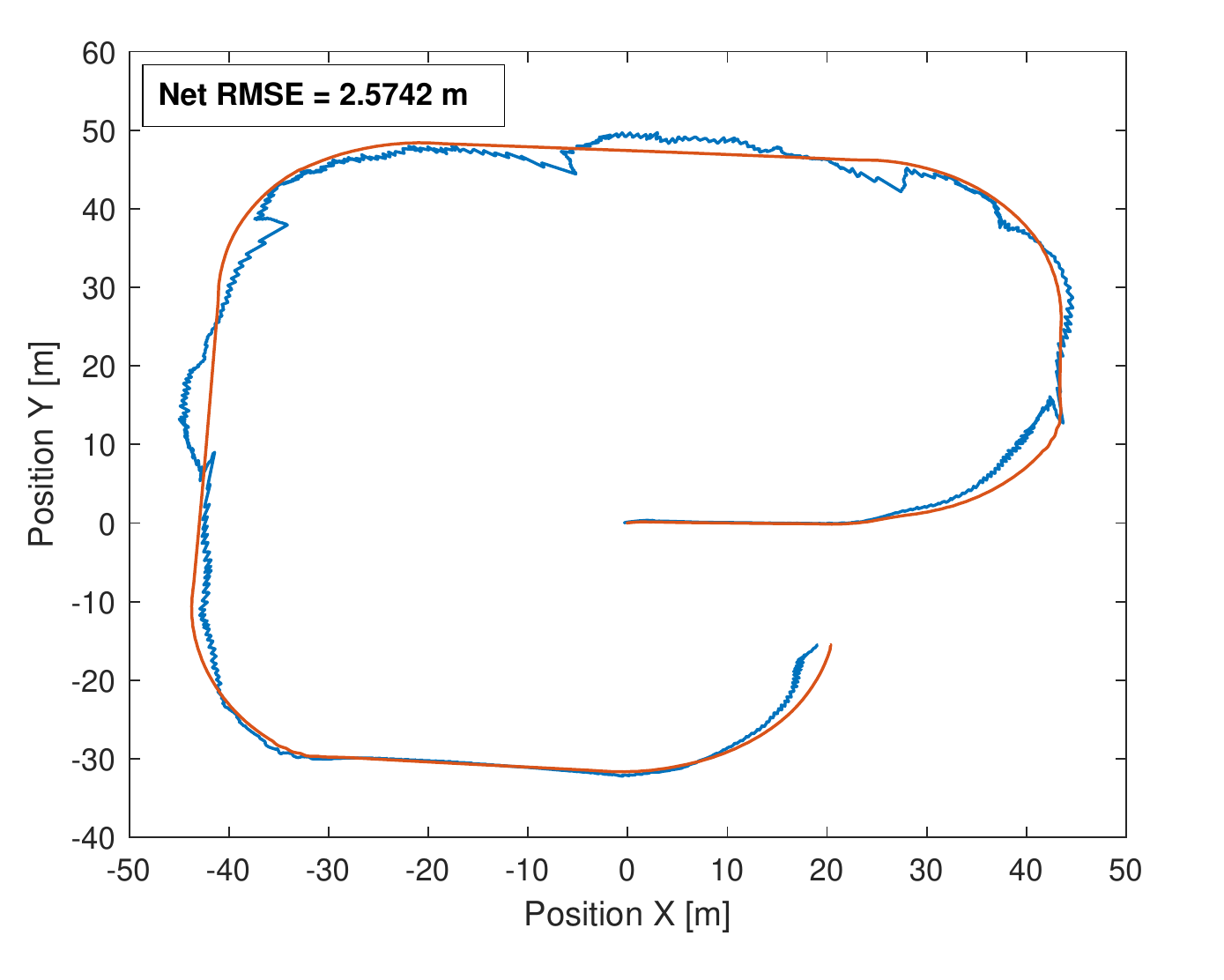}} 
            \hfil
        \subfloat[EKF with 1 IMU and 2 GPS]{
            \includegraphics[width=\w\textwidth]{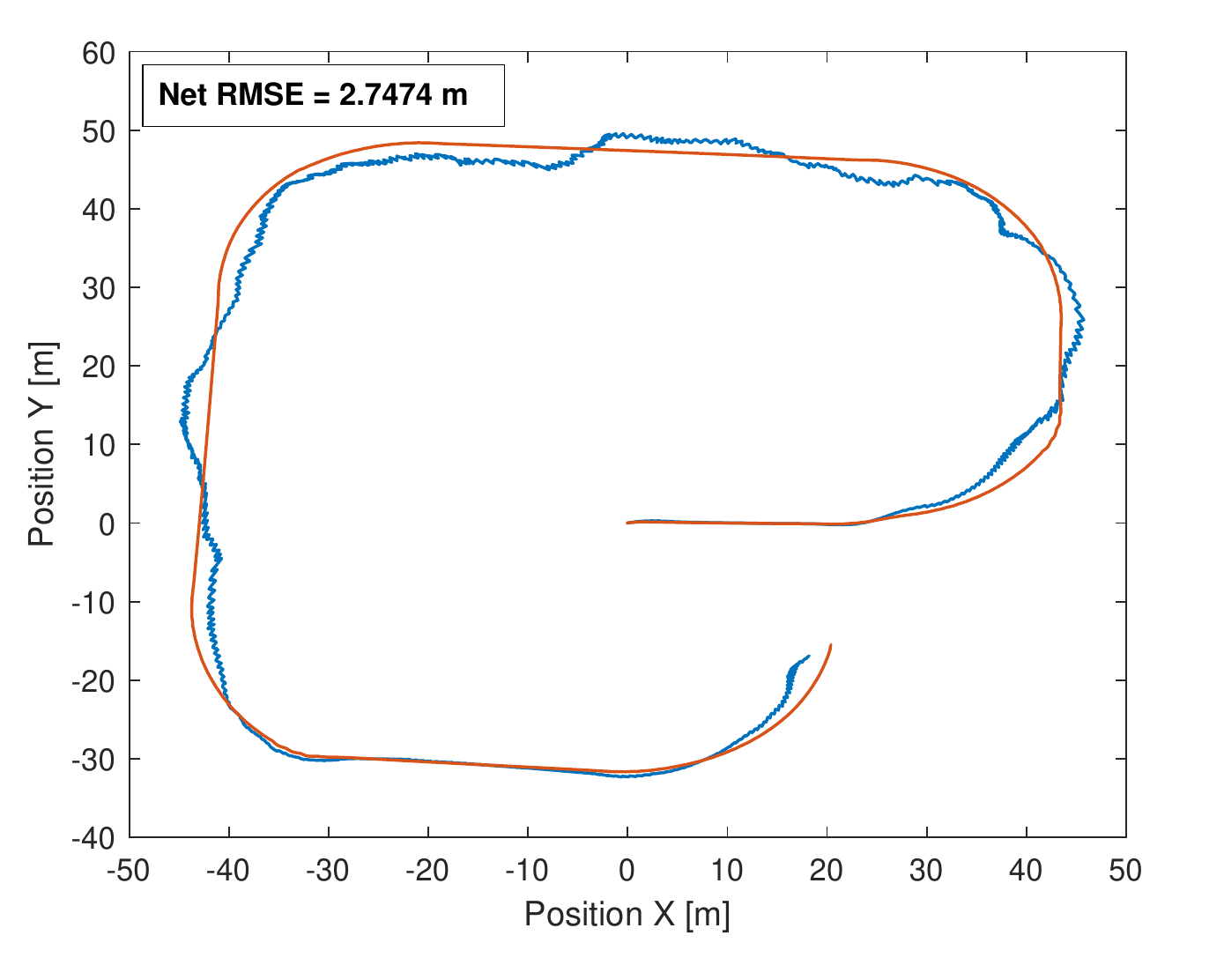}} 
            \hfil
        \subfloat[UKF with 1 IMU and 2 GPS]{
            \includegraphics[width=\w\textwidth]{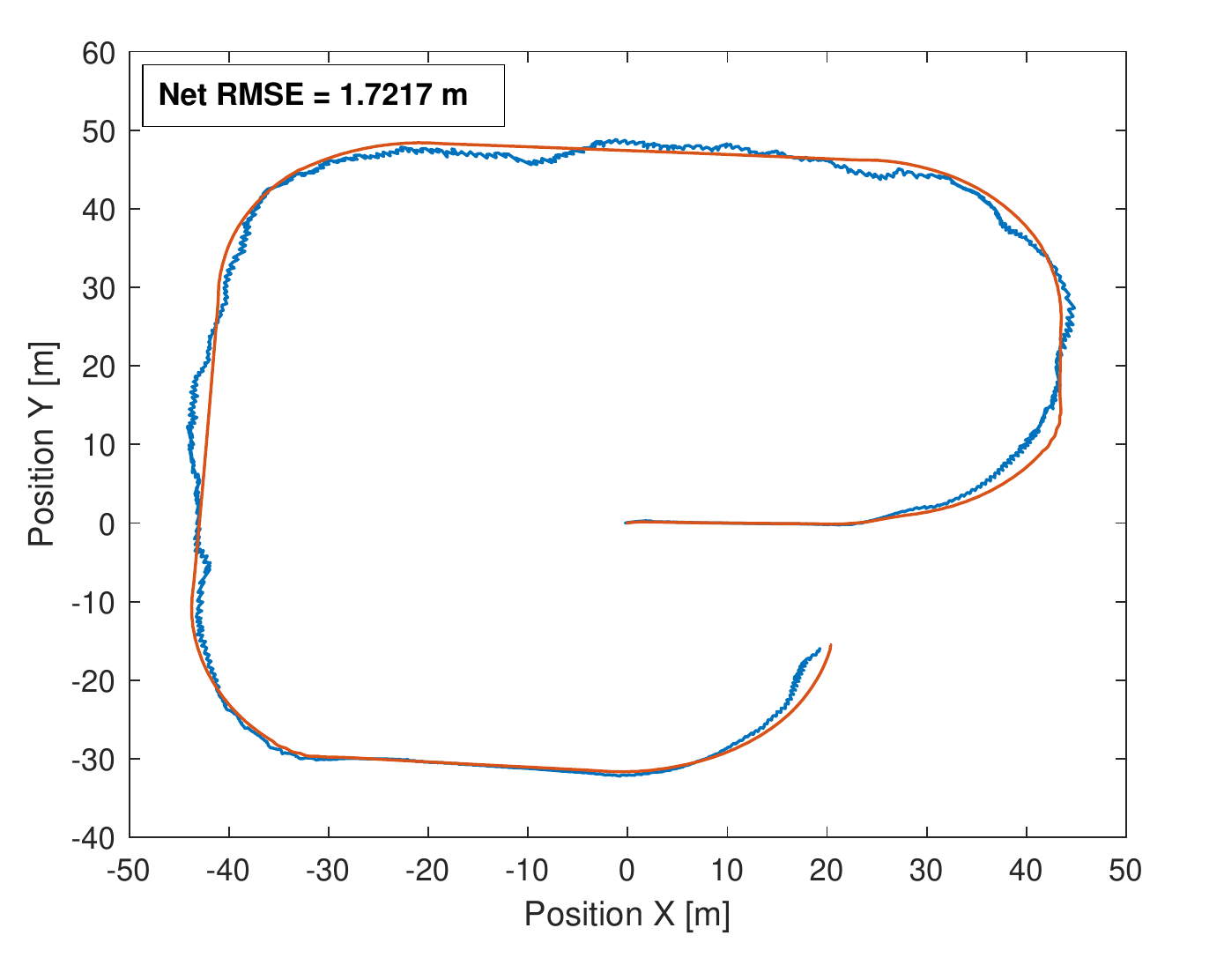}} 
            \hfil
        \subfloat[EKF with 2 IMU and 2 GPS]{
            \includegraphics[width=\w\textwidth]{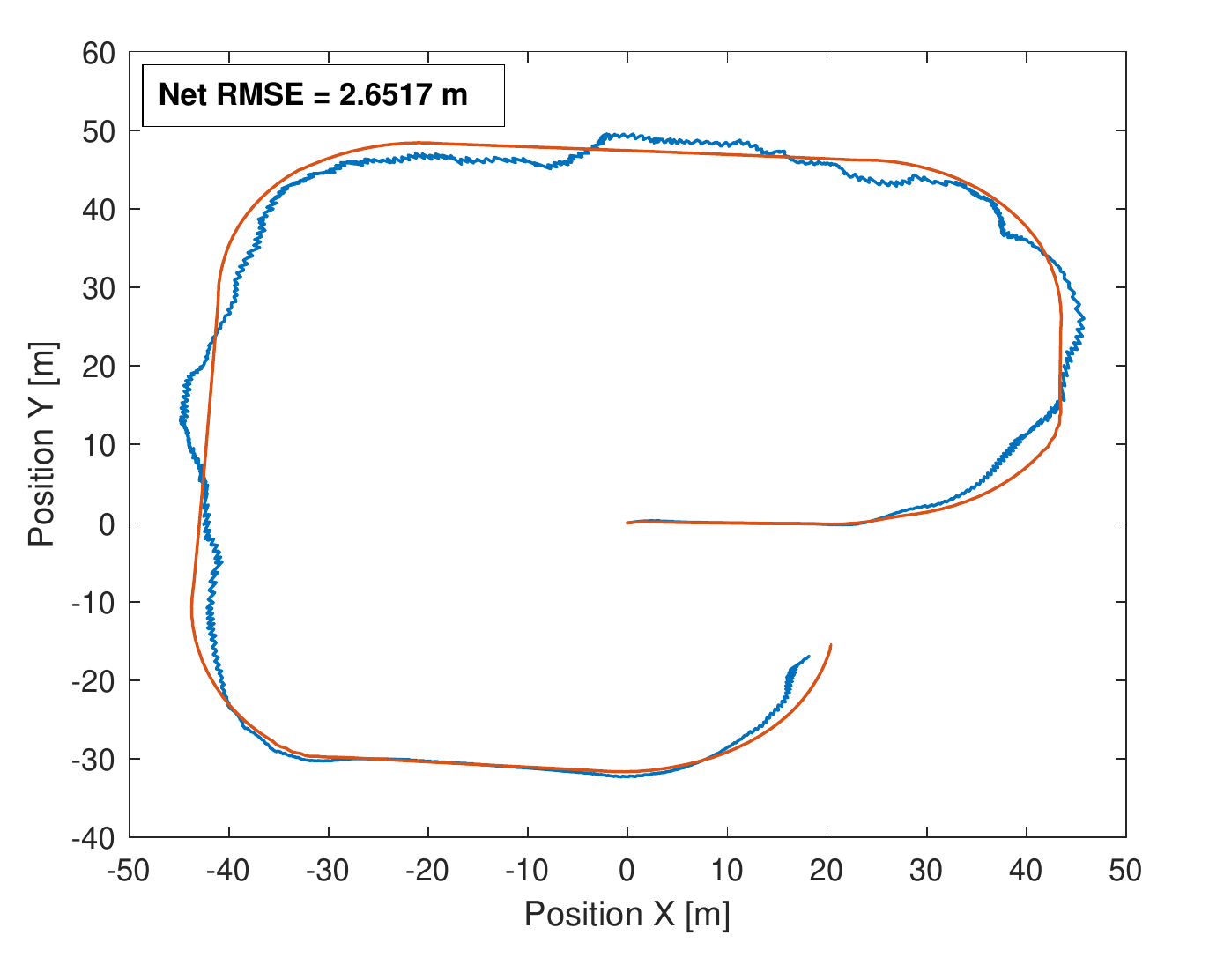}} 
            \hfil
        \subfloat[UKF with 2 IMU and 2 GPS]{
            \includegraphics[width=\w\textwidth]{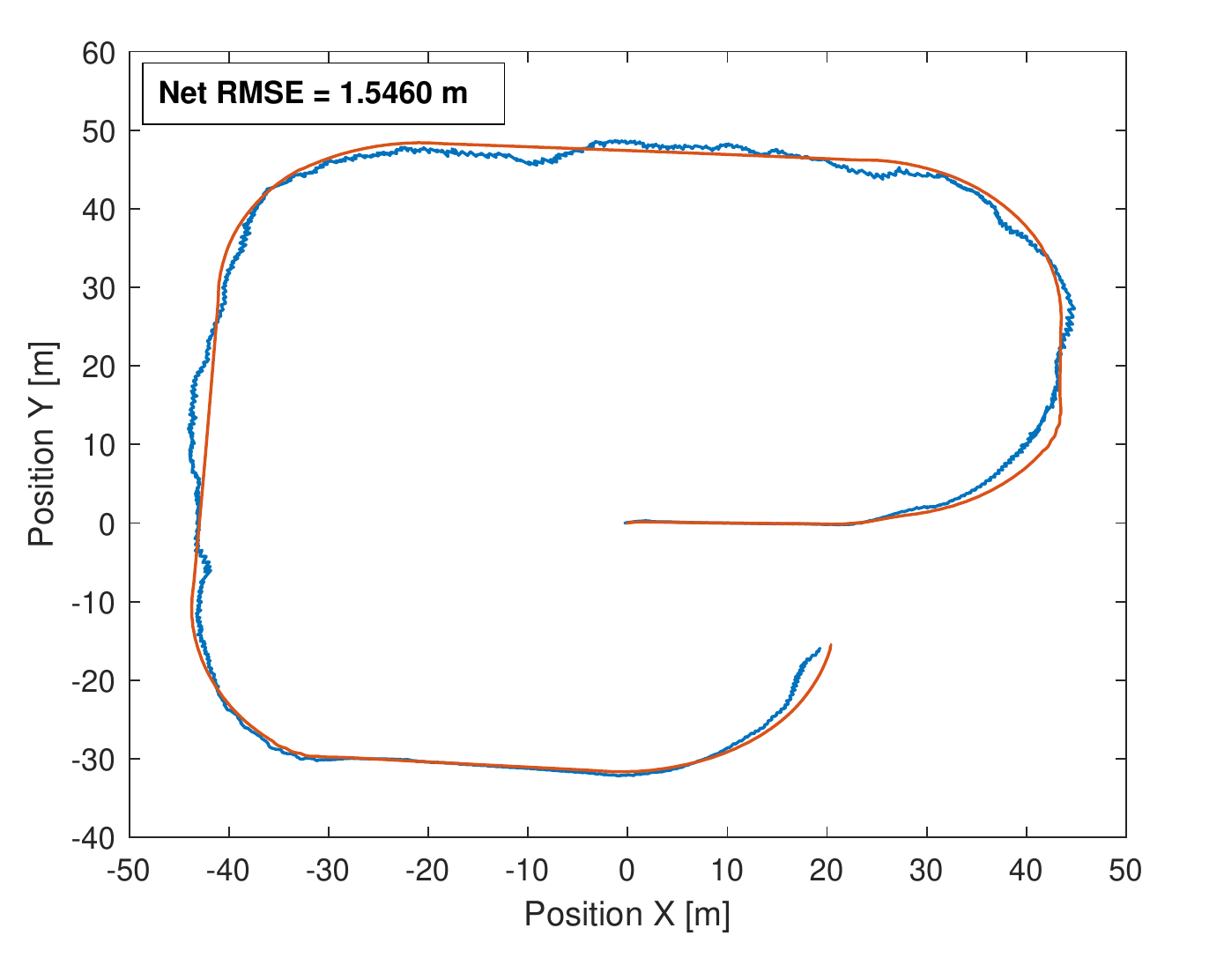}} 
            \hfil
        \subfloat[EKF with 3 IMU and 2 GPS]{
            \includegraphics[width=\w\textwidth]{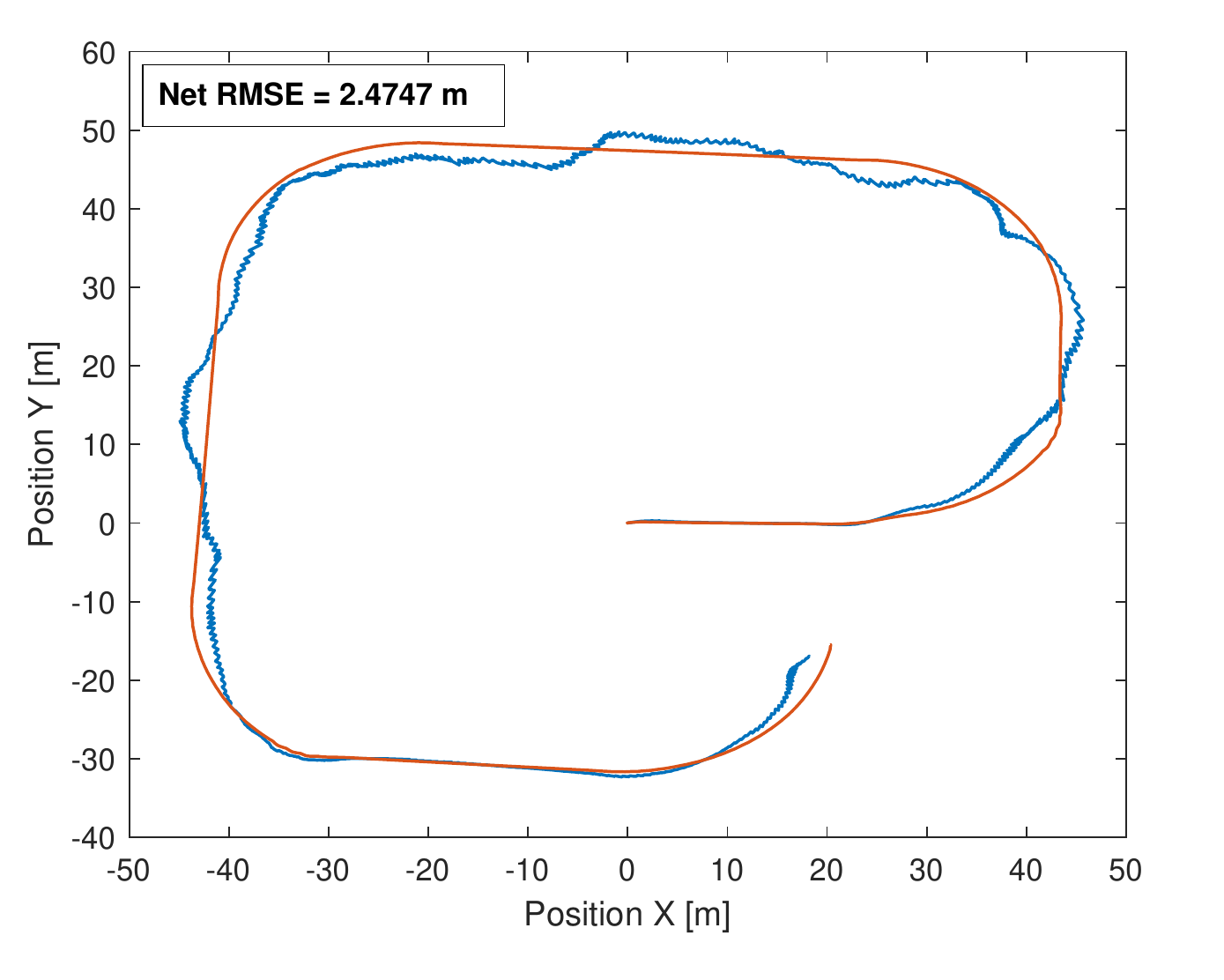}} 
            \hfil
        \subfloat[UKF with 3 IMU and 2 GPS]{
            \includegraphics[width=\w\textwidth]{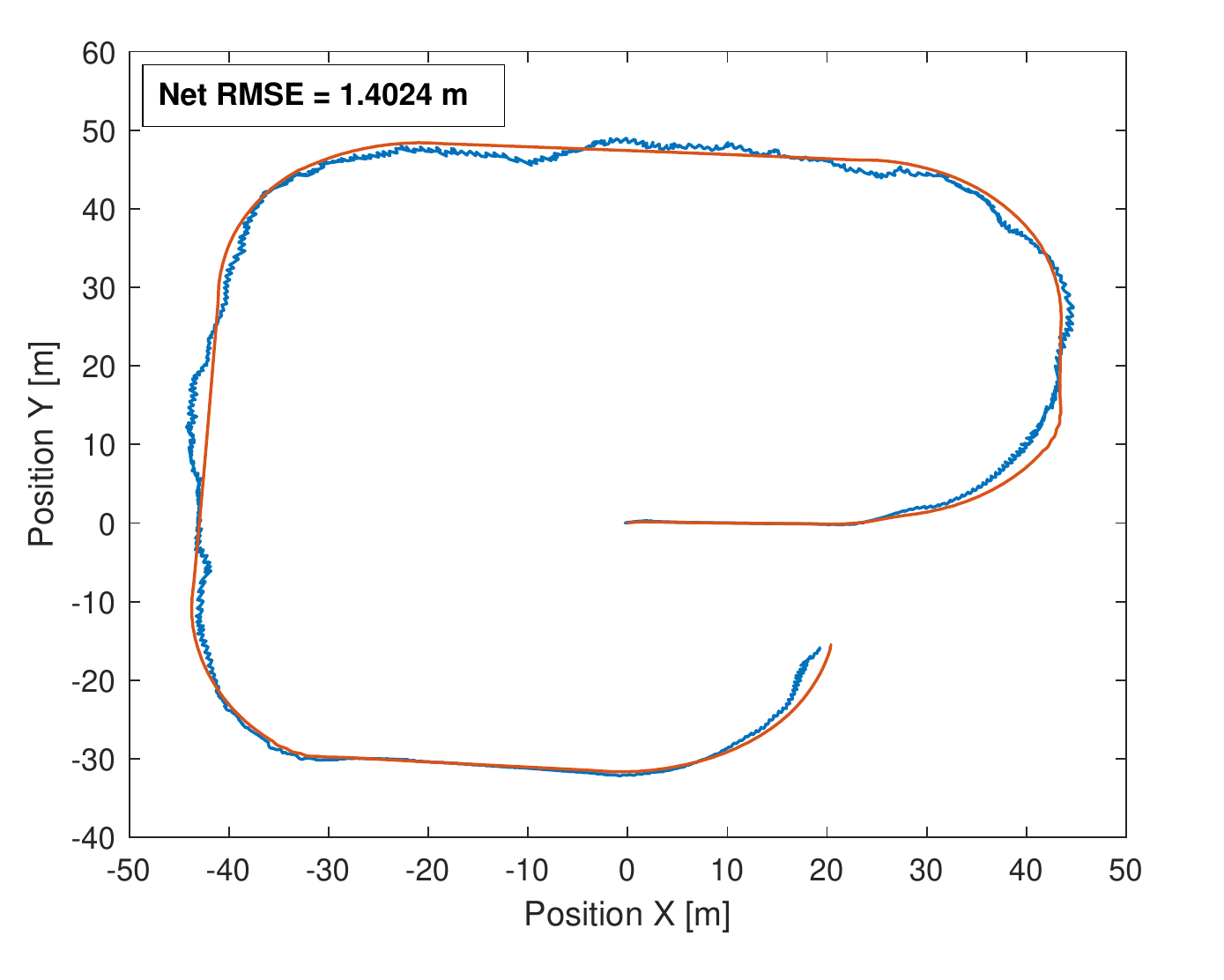}} 
            \hfil
        \subfloat[EKF with 3 IMU and 3 GPS]{
            \includegraphics[width=\w\textwidth]{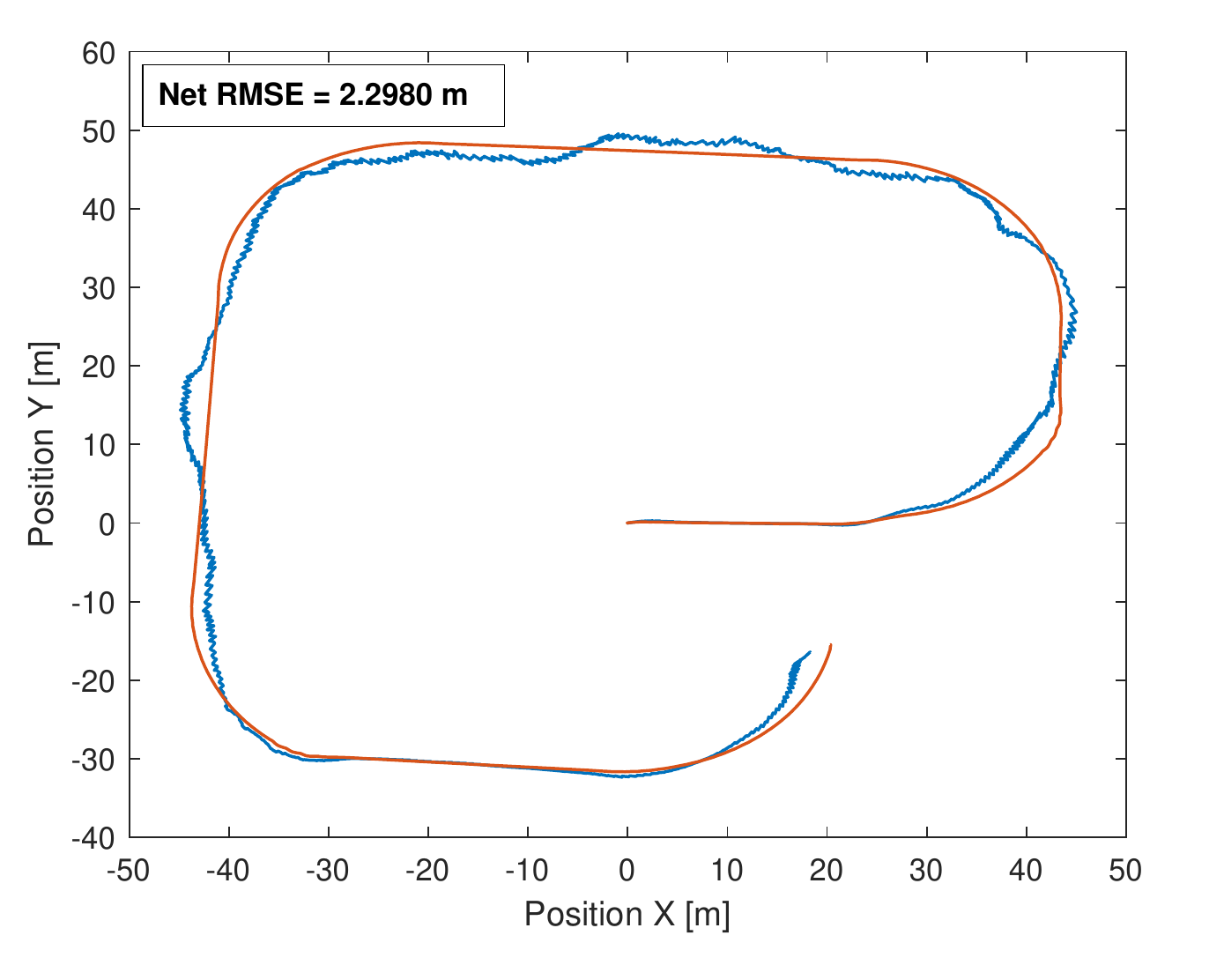}} 
            \hfil
        \subfloat[UKF with 3 IMU and 3 GPS]{
            \includegraphics[width=\w\textwidth]{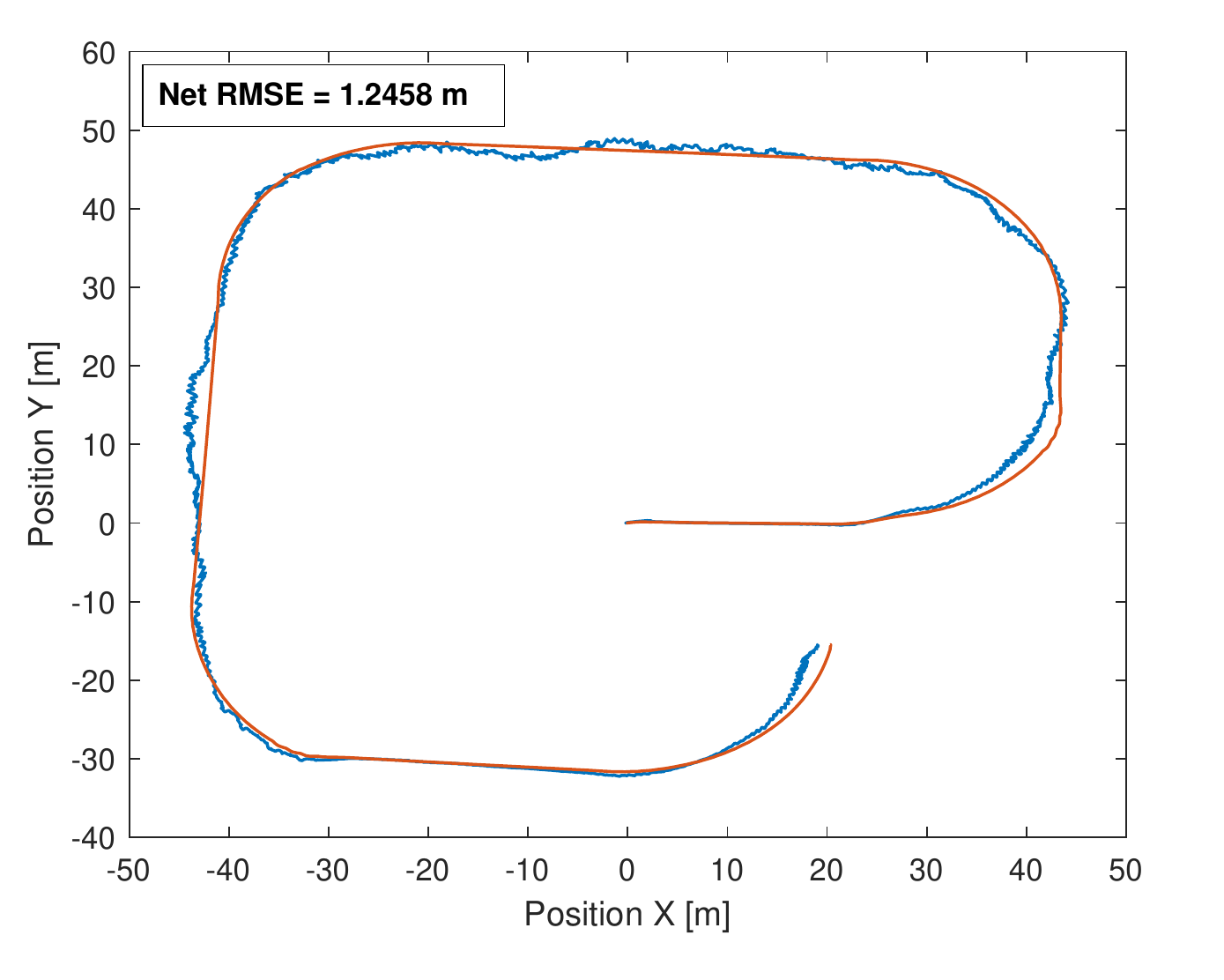}} 
            \hfil
        \caption{The IMU + Odometry estimation method yields for both EKF and UKF the RMSE over 70 meters.  Except for the IMU + Odometry methods, the other results can be divided into four performance levels according to the error scale. The worst level is EKF with one GPS, where the RMSE is about 3 $\sim $ 4 m. The second level is EKF with two GPS and EKF with three IMU and three GPS. In these cases, the RMSEs are about 2 $\sim $ 3 m. A better level is UKF with 1 GPS and 1 IMU, where the RMSE of the third level can be achieved about 2 $\sim $ 2.5 m. Finally, our experiment's best class is UKF with 2 GPS and 1 IMU, and UKF with 3 IMU and 3 GPS, which reduce the RMSE to about 1.2 m.} 
        \label{fig:overall results}
\end{figure*} 

Fig. \ref{fig:overall results} shows the wheel loader estimated trajectories given by different approaches and the ground truth trajectory, where the red lines are the ground truth trajectories that the vehicle passes, and the blue lines are the estimated position of the vehicle. Since only EKF and UKF are capable of handling the nonlinear problem, we compare the results obtained using EKF filtering and the UKF filtering technique with different sensor arrangements. Notice that, odometry sensor is always used though we do not explicitly mention it. Apparently, the UKF performs better than the EKF, which is also in line with the conclusion from most studies. Generally speaking, with GPS fusing in the estimation, the accuracy improves drastically. Also, since the GPS may lose signal every 10 seconds, an additional GPS sensor can surely increase the position accuracy. In contrast, more IMUs can only slightly improve the accuracy of positioning. As we can see, the IMU + Odometry estimation method yields inferior performance no matter with EKF or UKF. Once the IMU reports an inaccurate heading, the errors will be accumulated, causing the measured position drifts further away from its true position as the wheel loader travels.

\begin{figure*}[!t]
        \newcommand{\w}{0.45}
        \centering 
        \subfloat[RMSE of EKF ]{\includegraphics[width=\w\textwidth]{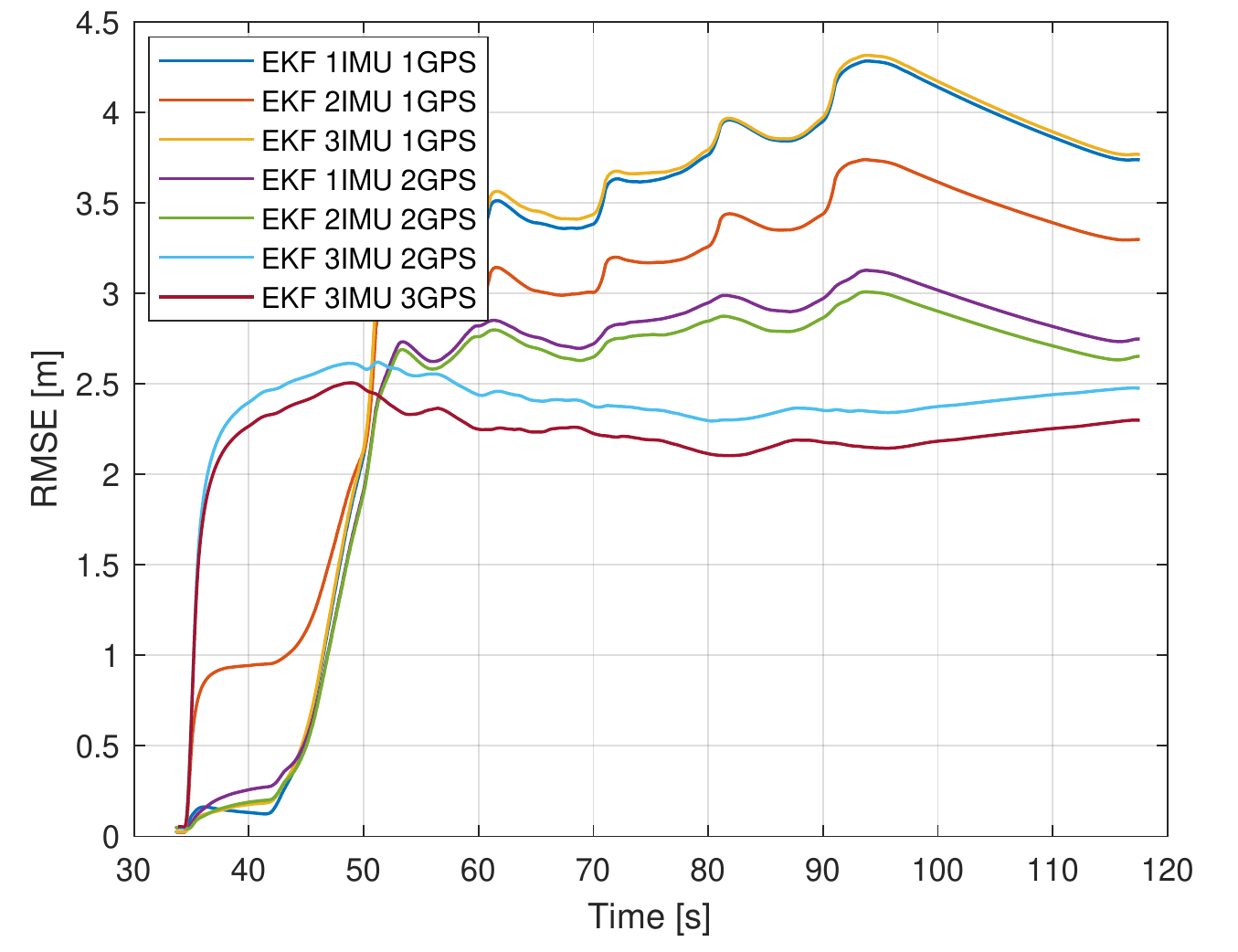}}
            \hfil 
        \subfloat[RMSE of UKF]{
            \includegraphics[width=\w\textwidth]{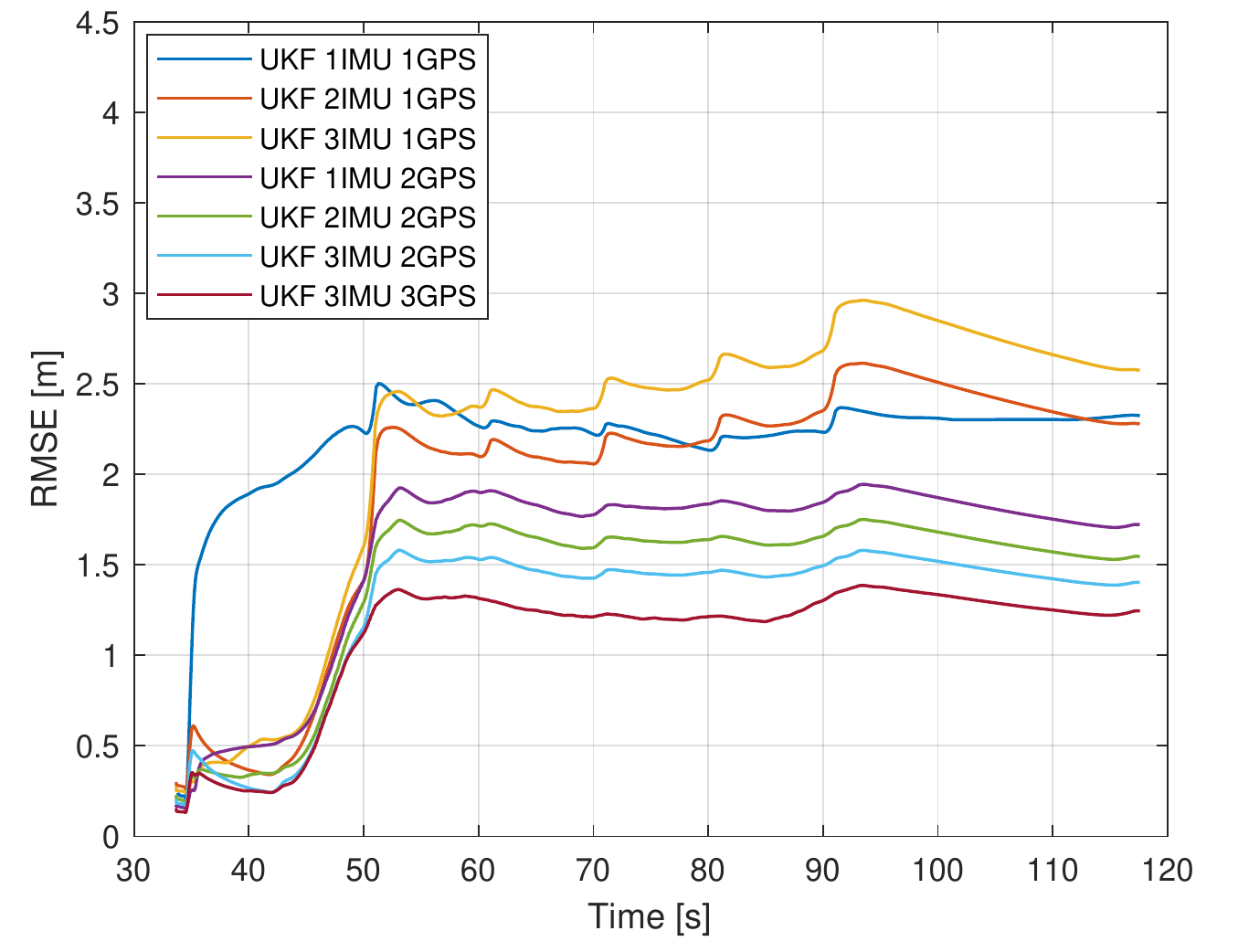}} 
            \hfil
        \subfloat[Euclidean distance error of EKF]{\includegraphics[width=\w\textwidth]{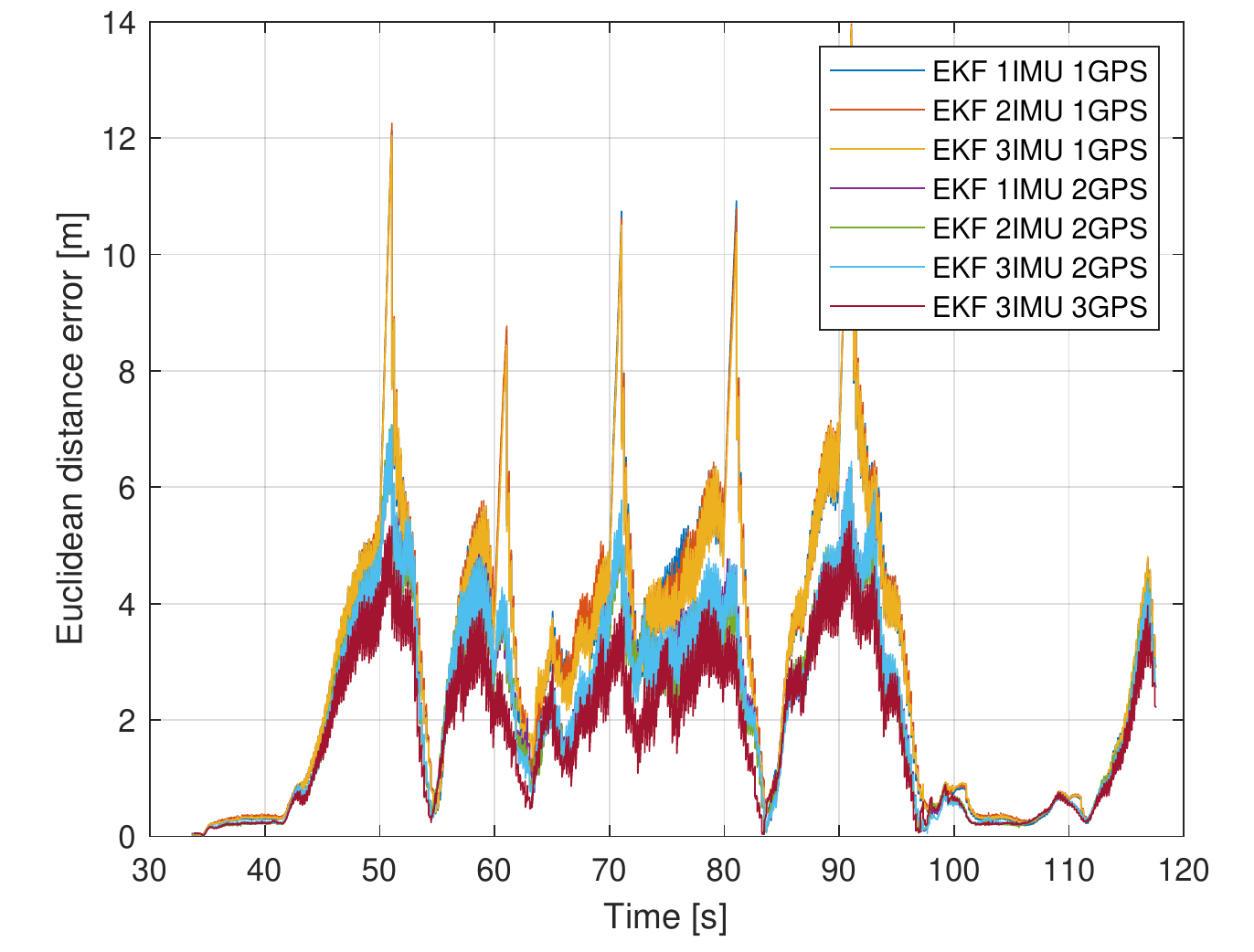}}
            \hfil 
        \subfloat[Euclidean distance error of UKF]{
            \includegraphics[width=\w\textwidth]{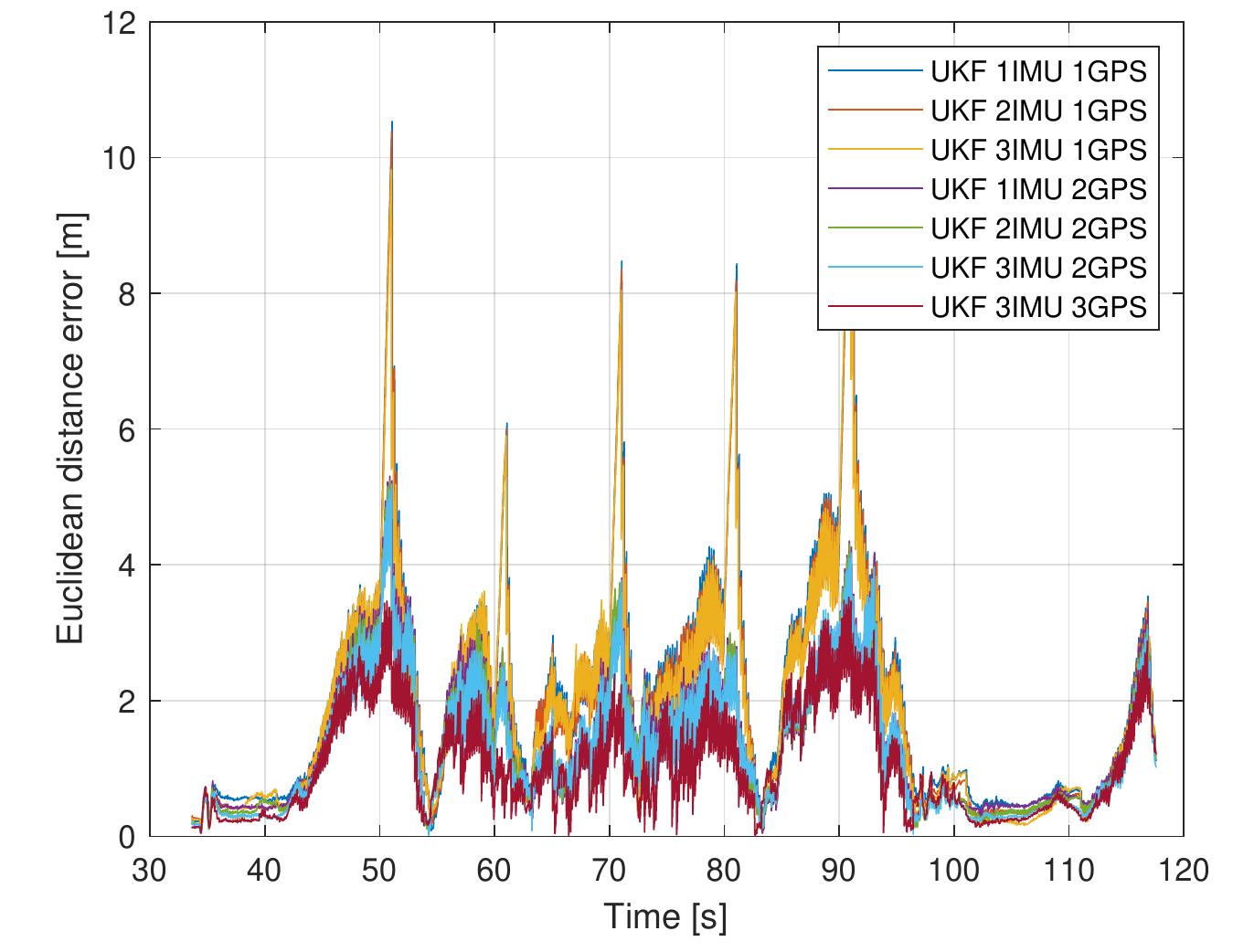}} 
            \hfil            
        \caption{Quantitative evaluation of different methods. Here (a) and (b) show the accumulative error, i.e., RMSE, while (c) and (d) demonstrate the current Euclidean distance error of each sensor arrangement. There are some noticeable instantaneous position changes every ten seconds due to infrequent GPS signal loss. Generally, UKF shows better performance than EKF for both RMSE and euclidean distance error.} 
        \label{fig:error comparsion}
\end{figure*} 

Fig. \ref{fig:error comparsion} shows a comparison of localization error between the different approaches with respect to time.
As shown in Fig. \ref{fig:error comparsion}, the RMSE and the euclidean distance error of each sensor arrangement can be obtained, indicating that the EKF has more significant tracking error than UKF.  It happens because the linearization through its Jacobian is an approximation, and the kinematic model of the wheel loader is highly nonlinear. To get a more intuitive and accurate description of the error for each sensor configuration and method, we create Tab. \ref{tab:Localization errors of each approach} to demonstrate the RMSE in detail. 

As aforementioned, the GPS signal will be lost for one second every ten seconds. Thus, there are some noticeable instantaneous position changes every ten seconds. The loss of GPS signals clearly causes these jumps. As we can see, for sensor configuration with only one GPS, the leap of the localization error as well as the variance values are more sharply. Obviously, with the number of GPS increases, the jumps are diminished and therefore become more acceptable.

\begin{table}
    \centering
    \caption{Localization errors of each approach }
    \scalebox{0.8}{
    \begin{tabular}{|c|c|c|}
       \hline
       \multirow{2}*{\emph{\textbf{Group}}} &  \emph{\textbf{RMSE}} \textbf{(m)}& \multirow{2}*{\emph{\textbf{Net RMSE}}\textbf{(m)}} \\
        ~ & {\textbf{(x,y)}} & ~ \\
       \hline 1 (EKF 1 IMU)       & (69.122, 30.039) & 75.3671 \\
       \hline 2 (EKF 1 IMU 1 GPS) &  (2.472, 2.802)  & 3.7363  \\
       \hline 3 (EKF 2 IMU 1 GPS) &  (2.172, 2.480)  & 3.2963 \\
       \hline 4 (EKF 3 IMU 1 GPS) &  (2.474, 2.840)  & 3.7660  \\
       \hline 5 (EKF 1 IMU 2 GPS) &  (1.830, 2.050)  & 2.7474  \\
       \hline 6 (EKF 2 IMU 2 GPS) &  (1.778, 1.968)  & 2.6517  \\
       \hline 7 (EKF 3 IMU 2 GPS) &  (1.894, 1.593)  & 2.4747  \\
       \hline 8 (EKF 3 IMU 3 GPS) &  (1.794, 1.437)  & 2.2980  \\
       \hline 9 (UKF 1 IMU)       & (56.586, 32.944) & 65.4774 \\
       \hline 10 (UKF 1 IMU 1 GPS) & (1.654, 1.632)  & 2.3234  \\
       \hline 11 (UKF 2 IMU 1 GPS) & (1.374, 1.817)  & 2.2781  \\
       \hline 12 (UKF 3 IMU 1 GPS) & (1.611, 2.001)  & 2.5742  \\
       \hline 13 (UKF 1 IMU 2 GPS) & (1.093, 1.330)  & 1.7217  \\
       \hline 14 (UKF 2 IMU 2 GPS) & (0.975, 1.200)  & 1.5460  \\
       \hline 15 (UKF 3 IMU 2 GPS) & (0.873, 1.098)  & 1.4024 \\
       \hline 16 (UKF 3 IMU 3 GPS) & (0.826, 0.933)  & 1.2458  \\
       \hline 
    \end{tabular}}
    \label{tab:Localization errors of each approach}
\end{table}

Apparently, the simulation results show that UKF is a better approach using data collected by onboard sensors of the wheel loader in gazebo environments. Intuitively, more sensors represent higher accuracy. However, the results show that an appropriate number of sensors can achieve acceptable accuracy at a lower cost. As we can see in Tab. \ref{tab:Localization errors of each approach}, the RMSE of UKF with 1 IMU 2 GPS is 1.7217 m, and UKF with 3 IMU and 2 GPS is 1.4024 m. With additional 2 IMU and 1 GPS, the RMSE is only slightly reduced by 0.3 m. More importantly, the maximal error is strongly diminished by one additional GPS, whereas continually increasing the sensor number does not further reduce the error proportionally. Of course, according to different application scenarios, different sensor configurations shall be chosen. For our application scenario, UKF with 1 IMU and 2 GPS has sufficient accuracy and a better economy respecting sensor hardware cost and onboard ECU computational effort. Thus, we suggest using this sensor configuration to locate the mobile machines and then develop the realtime map plotter.

\subsection{Plotter results}

\begin{figure*}[!t]
        \newcommand{\w}{0.22}
        \centering 
        \subfloat[Predefined five areas of resistance plotted by OpenCV]{\includegraphics[width=\w\textwidth]{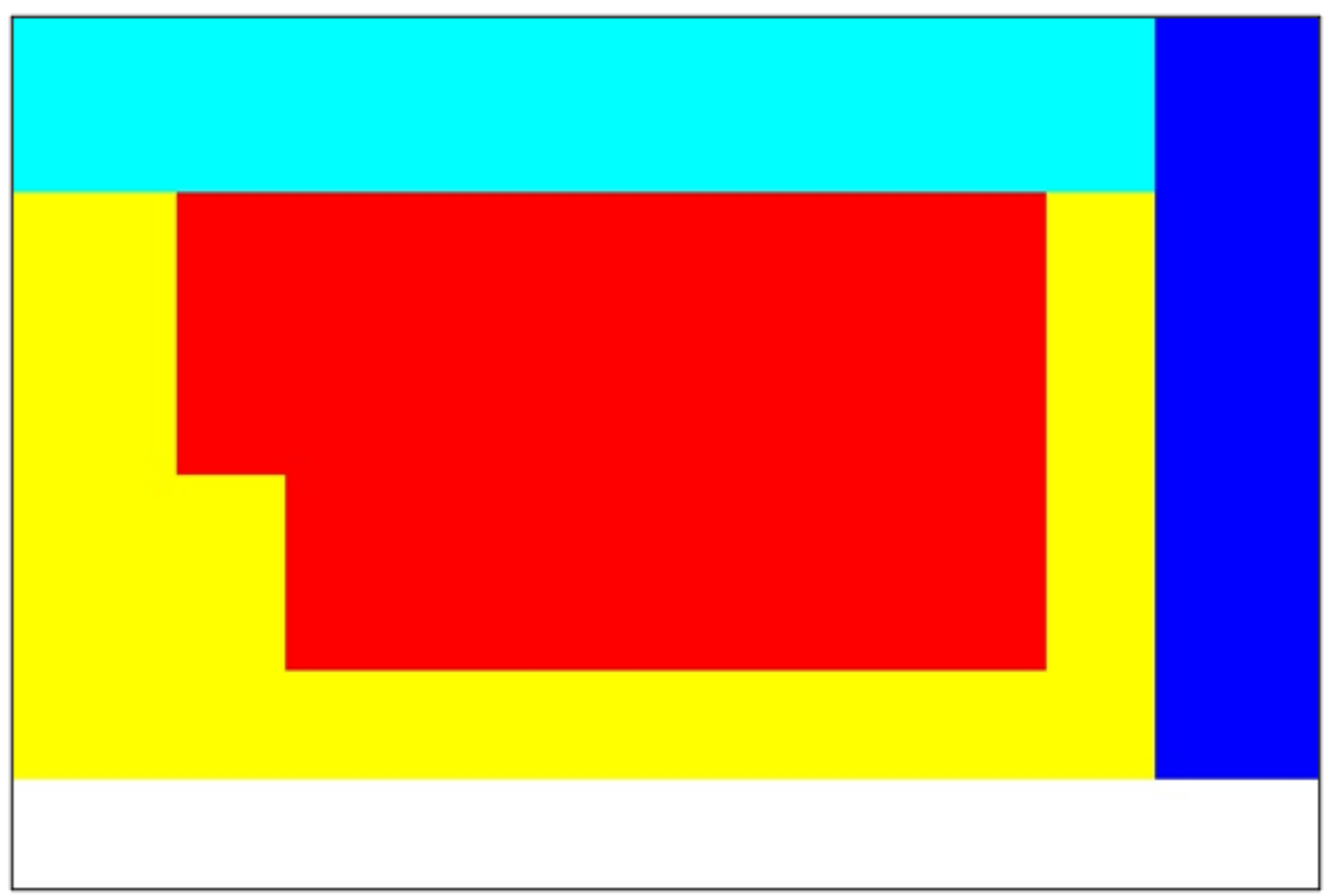}}
            \hfil 
        \subfloat[Result of ground resistance map with wheel loader (EKF with 1 IMU and 1 GPS)]{
            \includegraphics[width=\w\textwidth]{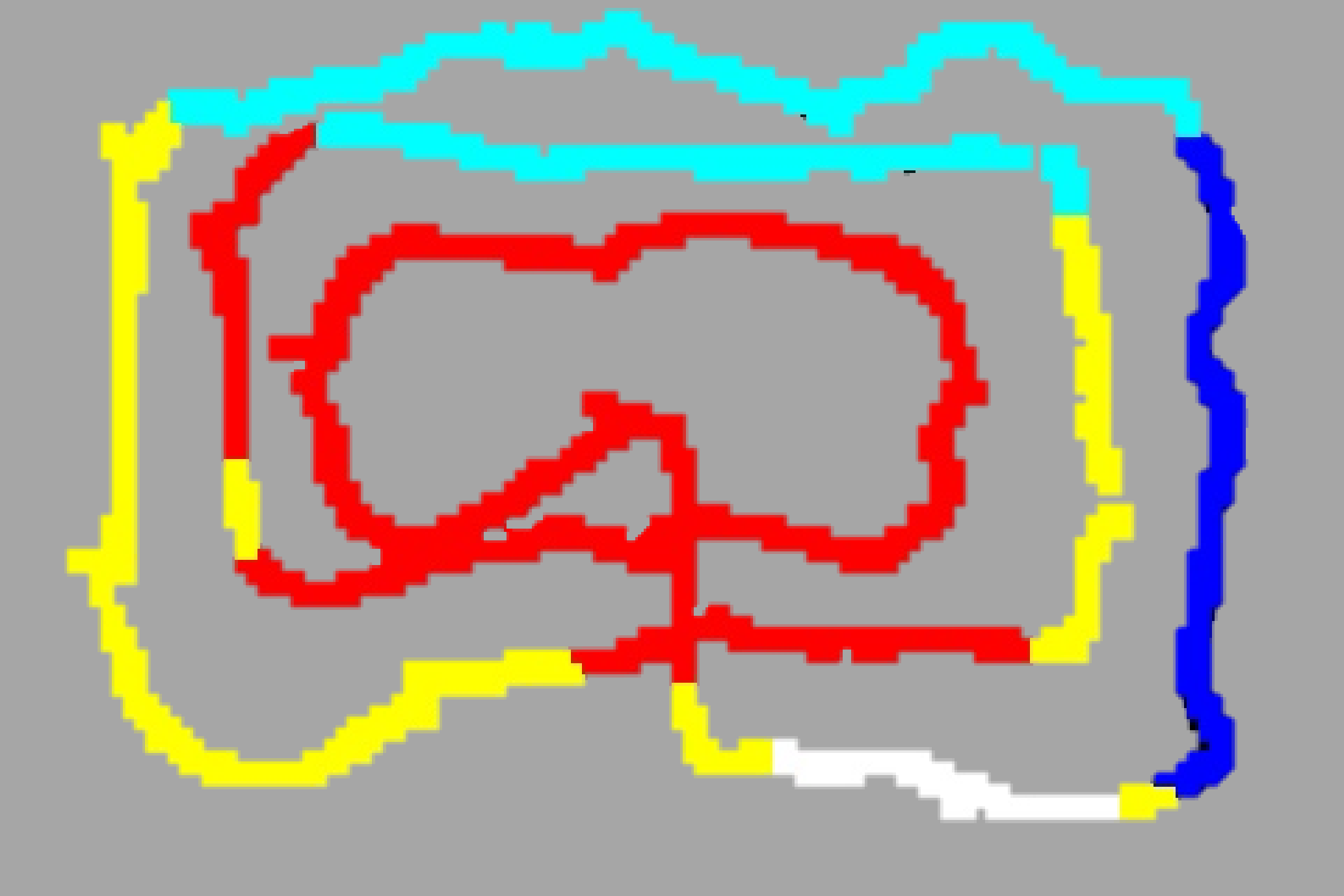}} 
            \hfil
        \subfloat[Result of ground resistance map with wheel loader (UKF with 1 IMU and 1 GPS)]{
            \includegraphics[width=\w\textwidth]{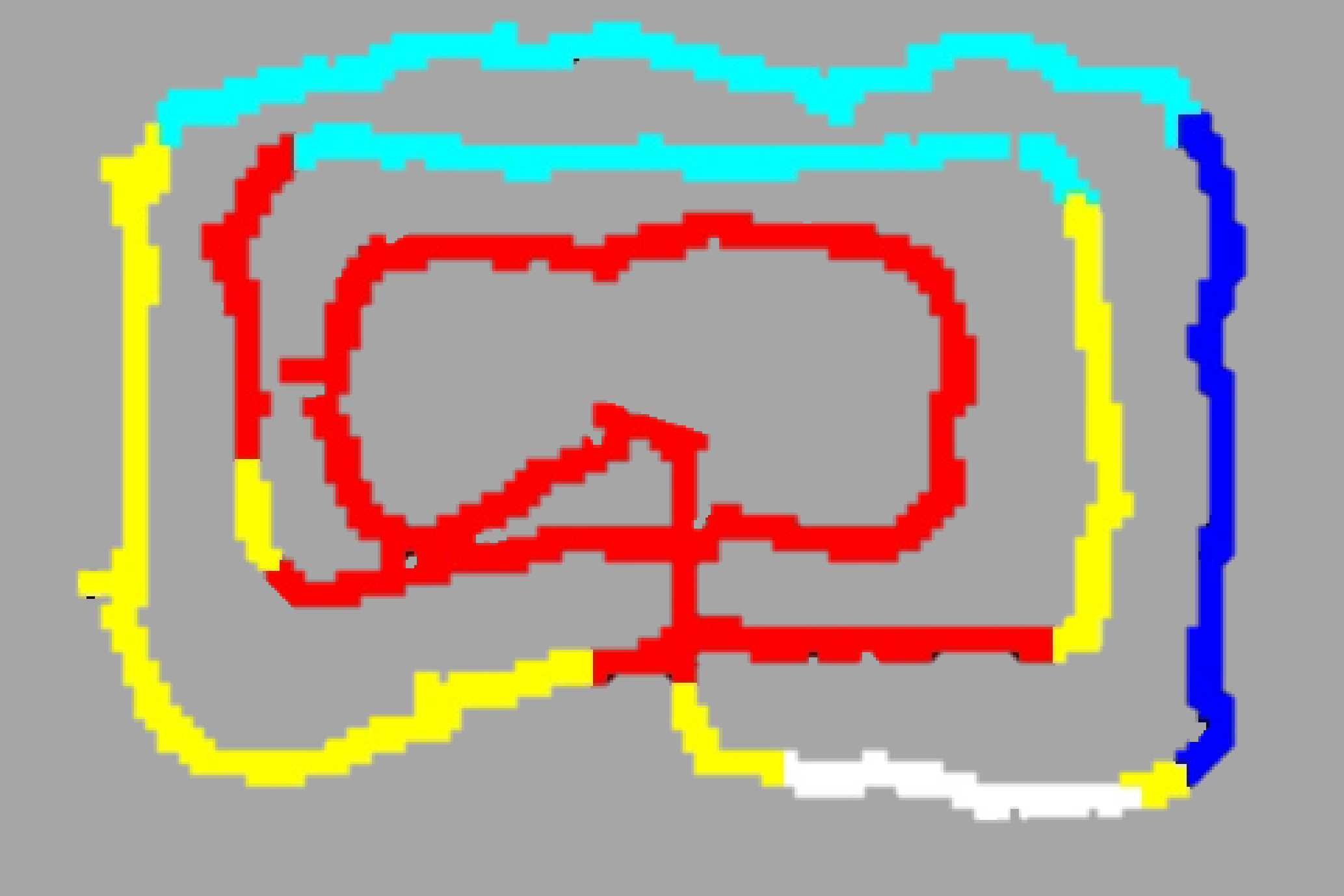}} 
            \hfil
        \subfloat[Result of ground resistance map with wheel loader (UKF with 1 IMU and 2 GPS)]{
            \includegraphics[width=\w\textwidth]{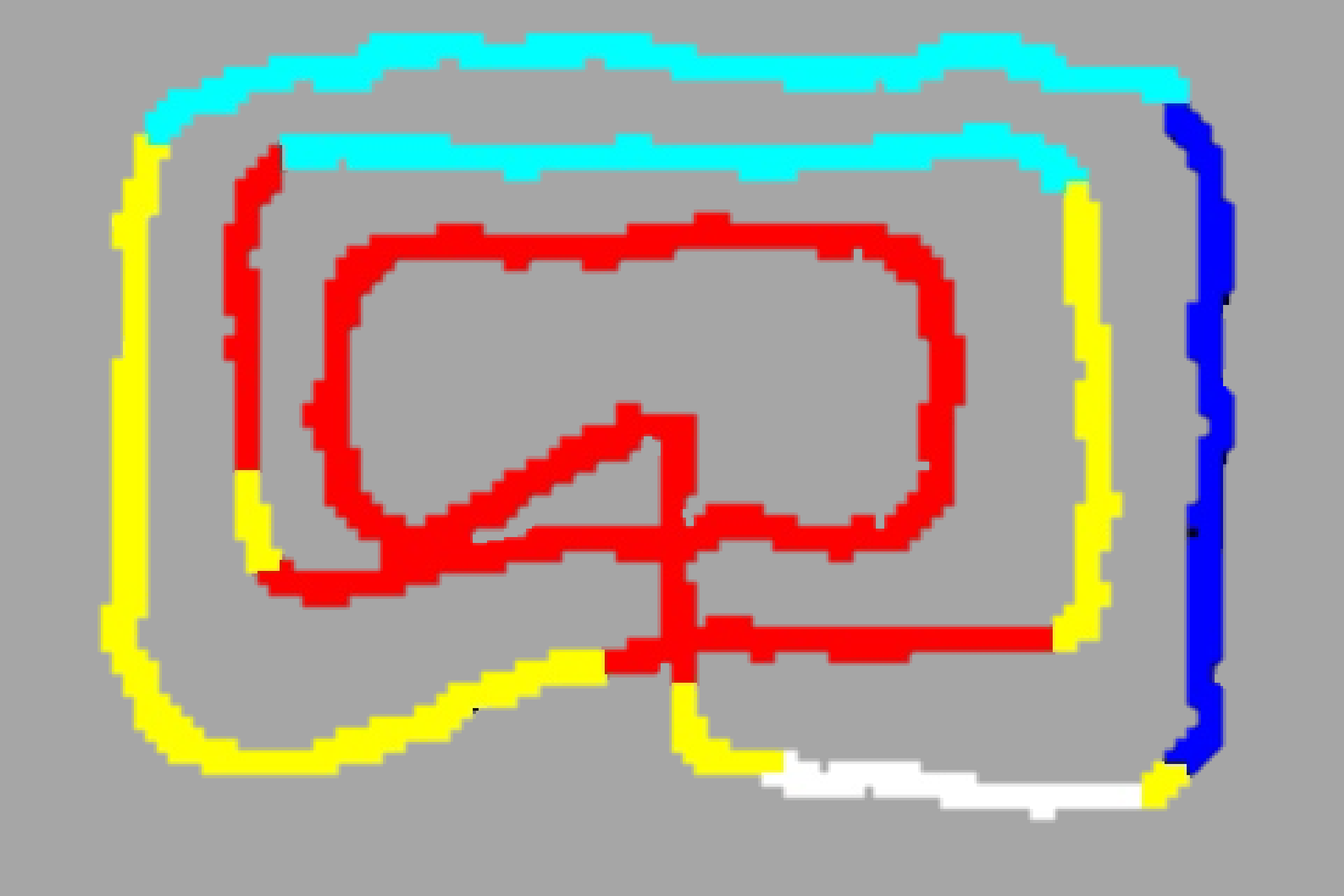}}
            \hfil 
        \subfloat[Predefined two areas of road grade plotted by OpenCV]{
            \includegraphics[width=\w\textwidth]{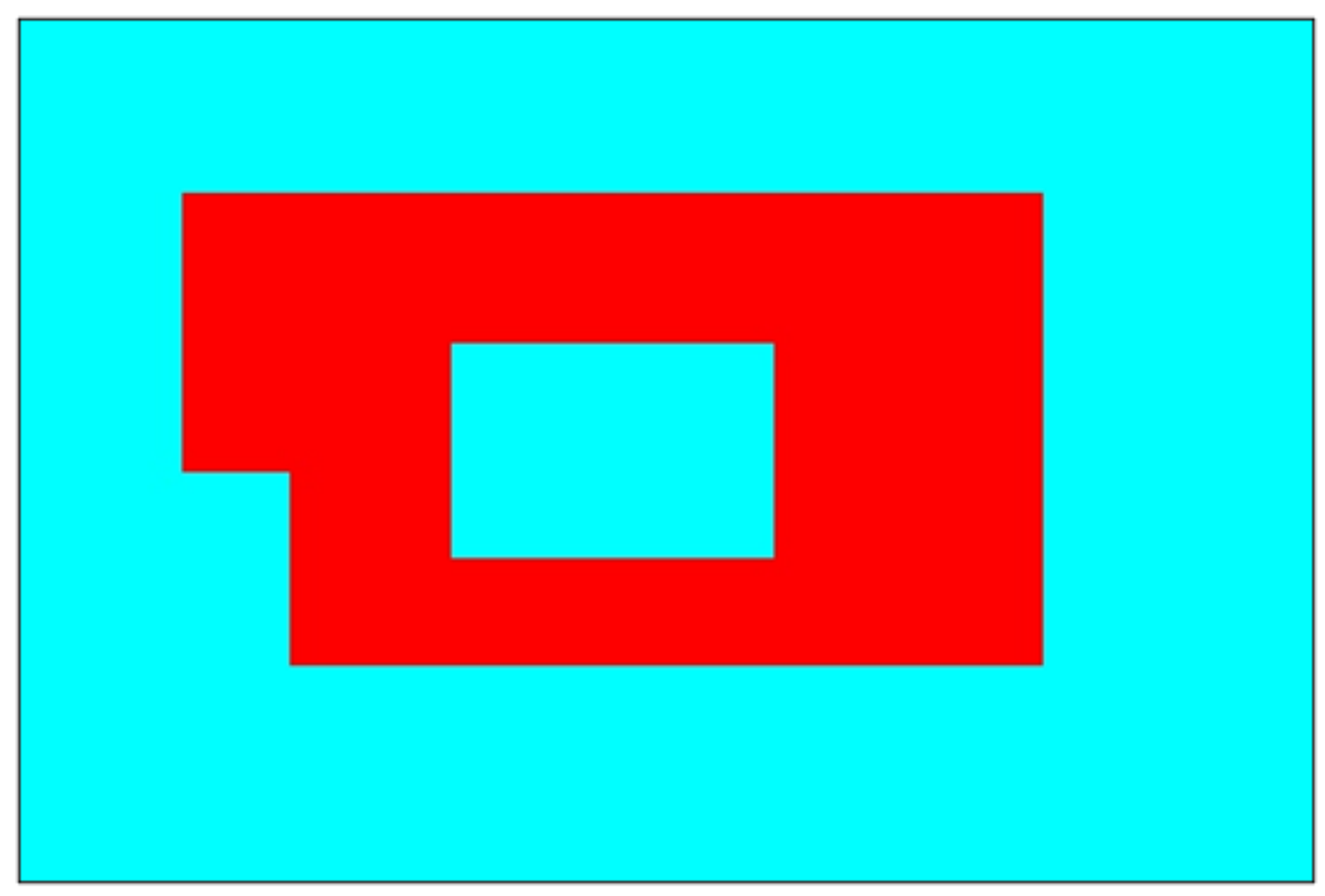}} 
            \hfil
        \subfloat[Result of road grade map with wheel loader (EKF with 1 IMU and 1 GPS)]{
            \includegraphics[width=\w\textwidth]{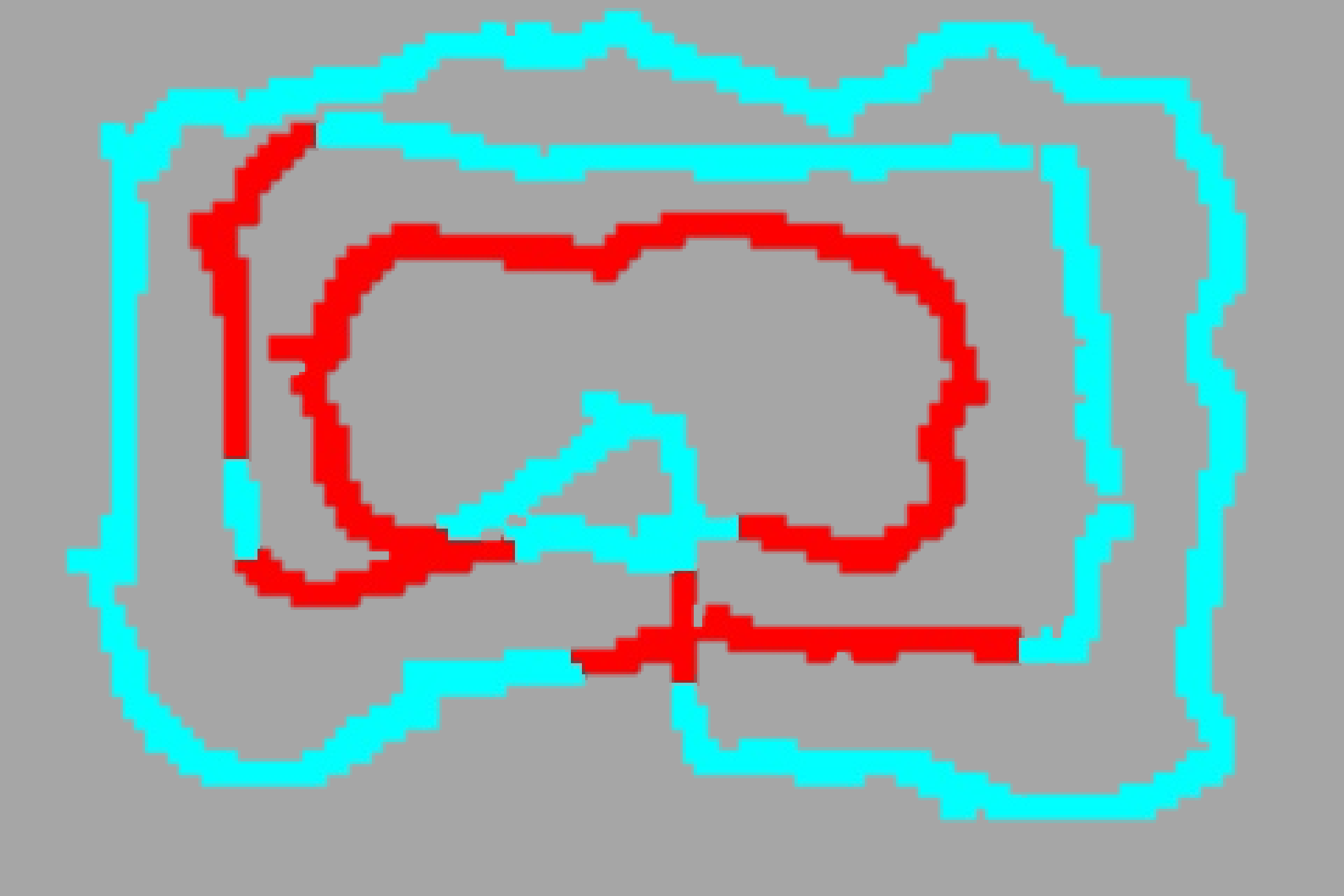}} 
            \hfil
        \subfloat[Result of road grade map with wheel loader (UKF with 1 IMU and 1 GPS)]{
            \includegraphics[width=\w\textwidth]{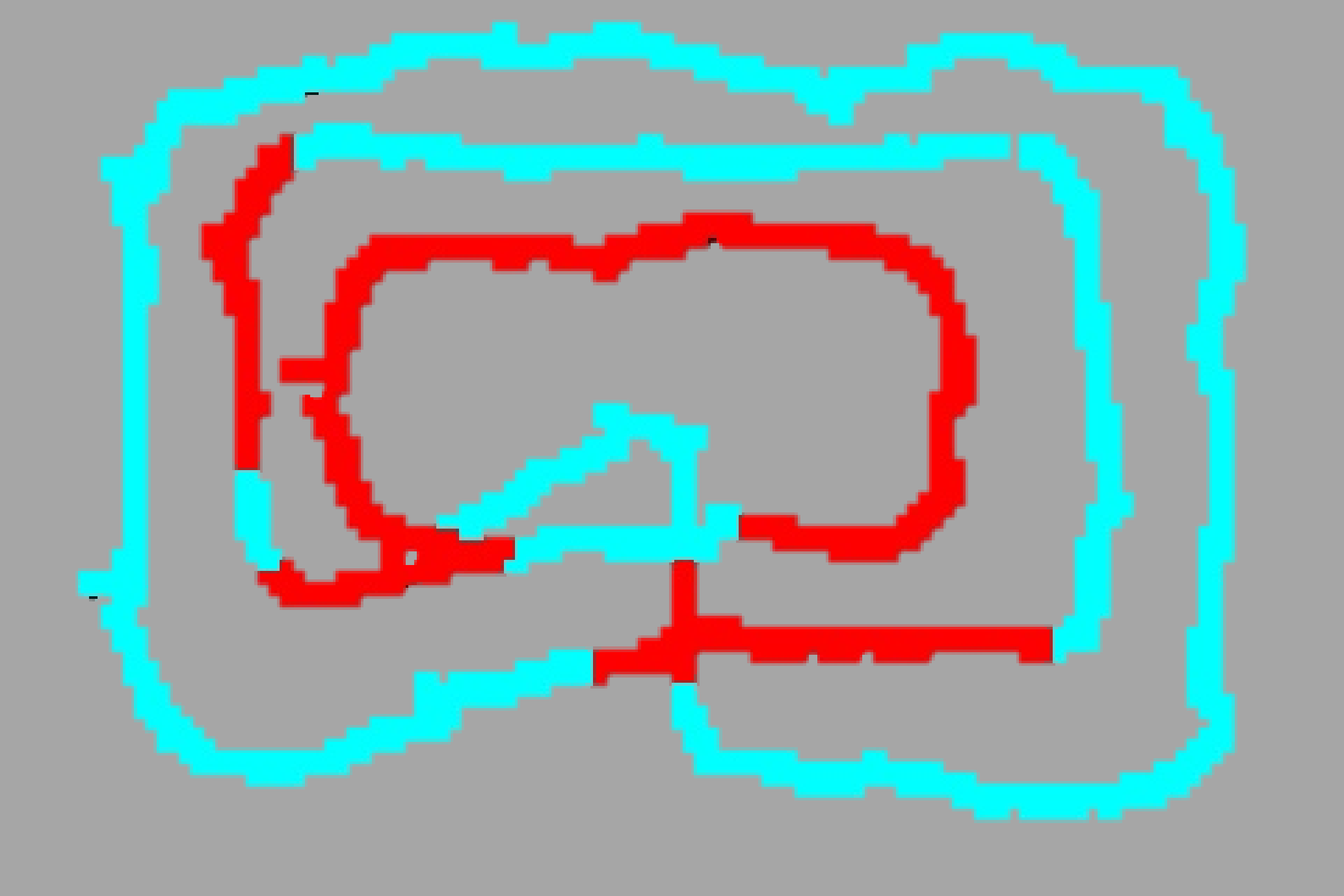}}
            \hfil 
        \subfloat[Result of road grade map with wheel loader (UKF with 1 IMU and 2 GPS)]{
            \includegraphics[width=\w\textwidth]{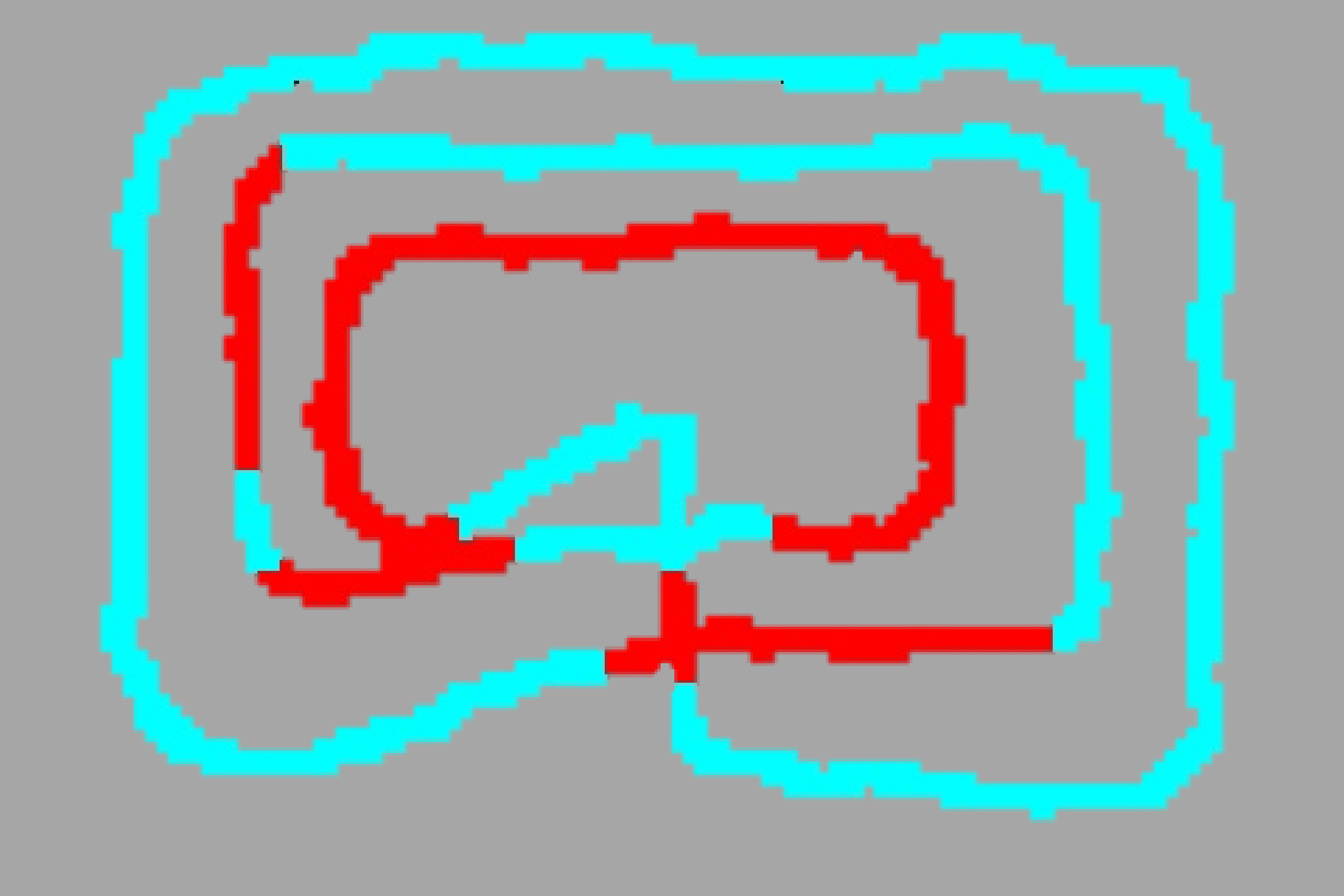}} 
            \hfil
        \caption{The ground truth and estimated maps. In the ground resistance map and road grade map plotted by EKF with 1 IMU and 1 GPS, the spikes caused by infrequent GPS are quite obvious. With additional GPS sensor fused in Kalman filter, the spikes improve a lot.} \label{fig:After we drive the wheel loader around the simulation plane, the results of the plotter were created}
\end{figure*}



\begin{figure*}[!t]
        \newcommand{\w}{0.33}
        \centering 
        \subfloat[Result of EKF with 1 IMU and 1 GPS]{\includegraphics[width=\w\textwidth]{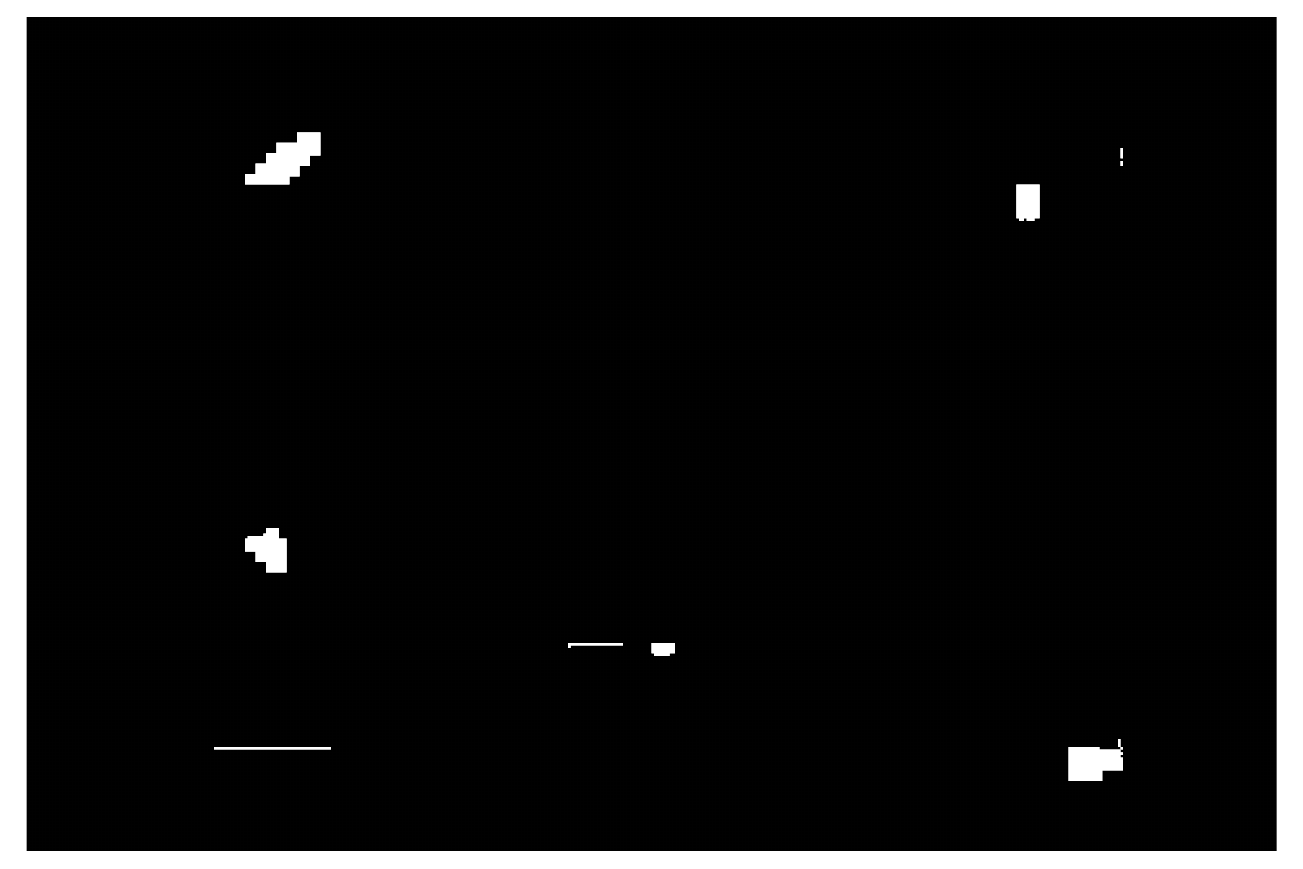}}
            \hfil 
        \subfloat[Result of UKF with 1 IMU and 1 GPS]{\includegraphics[width=\w\textwidth]{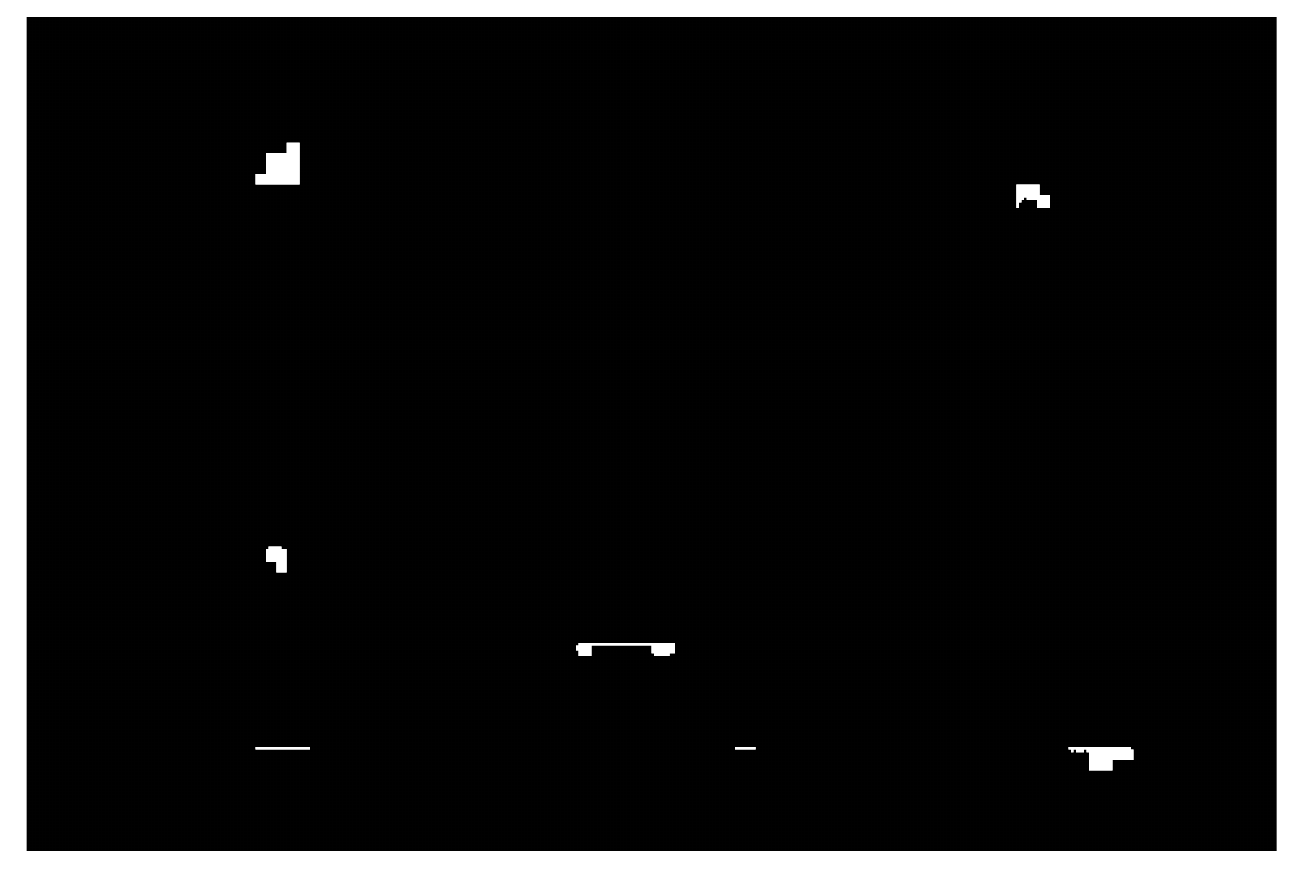}}
            \hfil 
        \subfloat[Result of UKF with 1 IMU and 2 GPS]{\includegraphics[width=\w\textwidth]{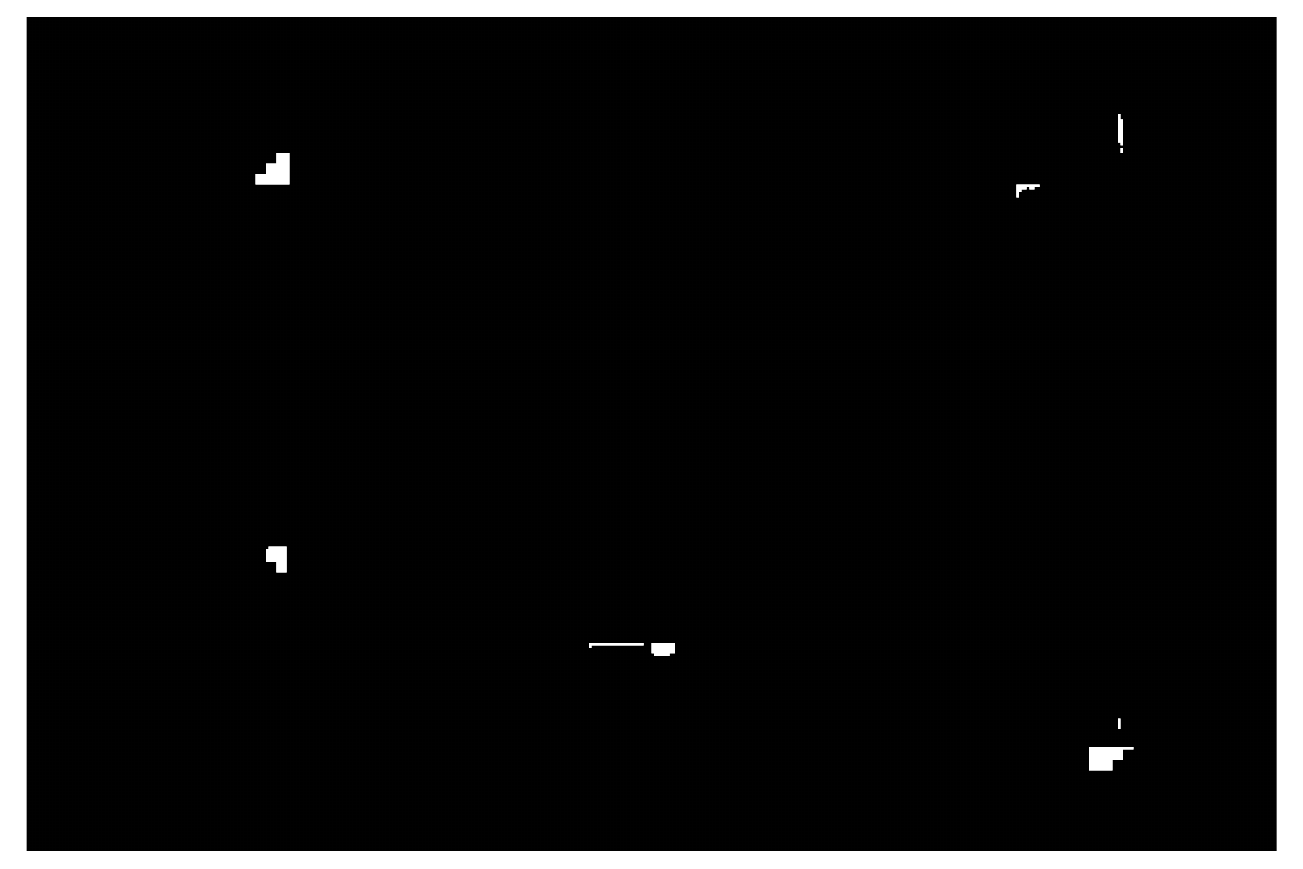}}
            \hfil   
        \subfloat[Result of EKF with 1 IMU and 1 GPS]{\includegraphics[width=\w\textwidth]{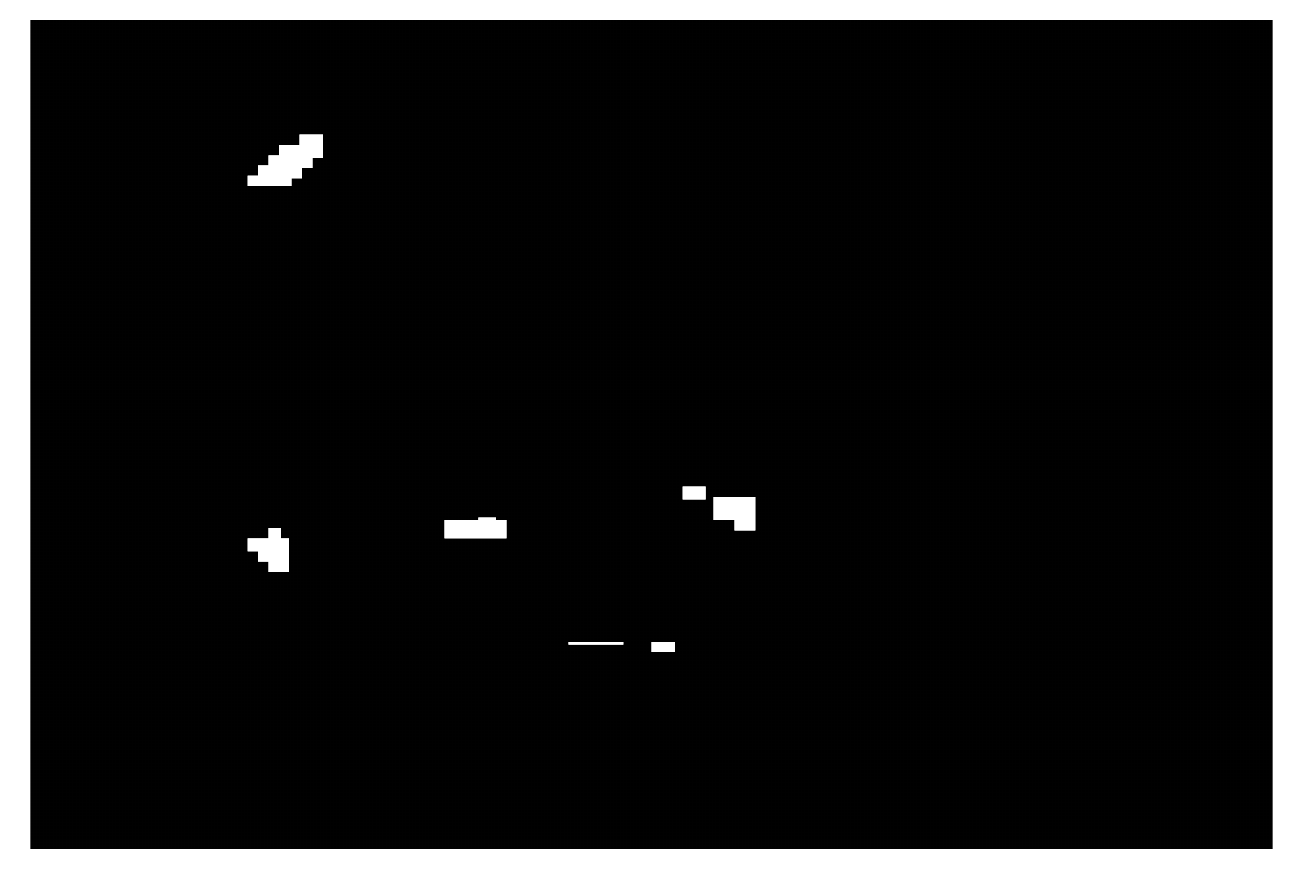}}
            \hfil            
        \subfloat[Result of UKF with 1 IMU and 1 GPS]{\includegraphics[width=\w\textwidth]{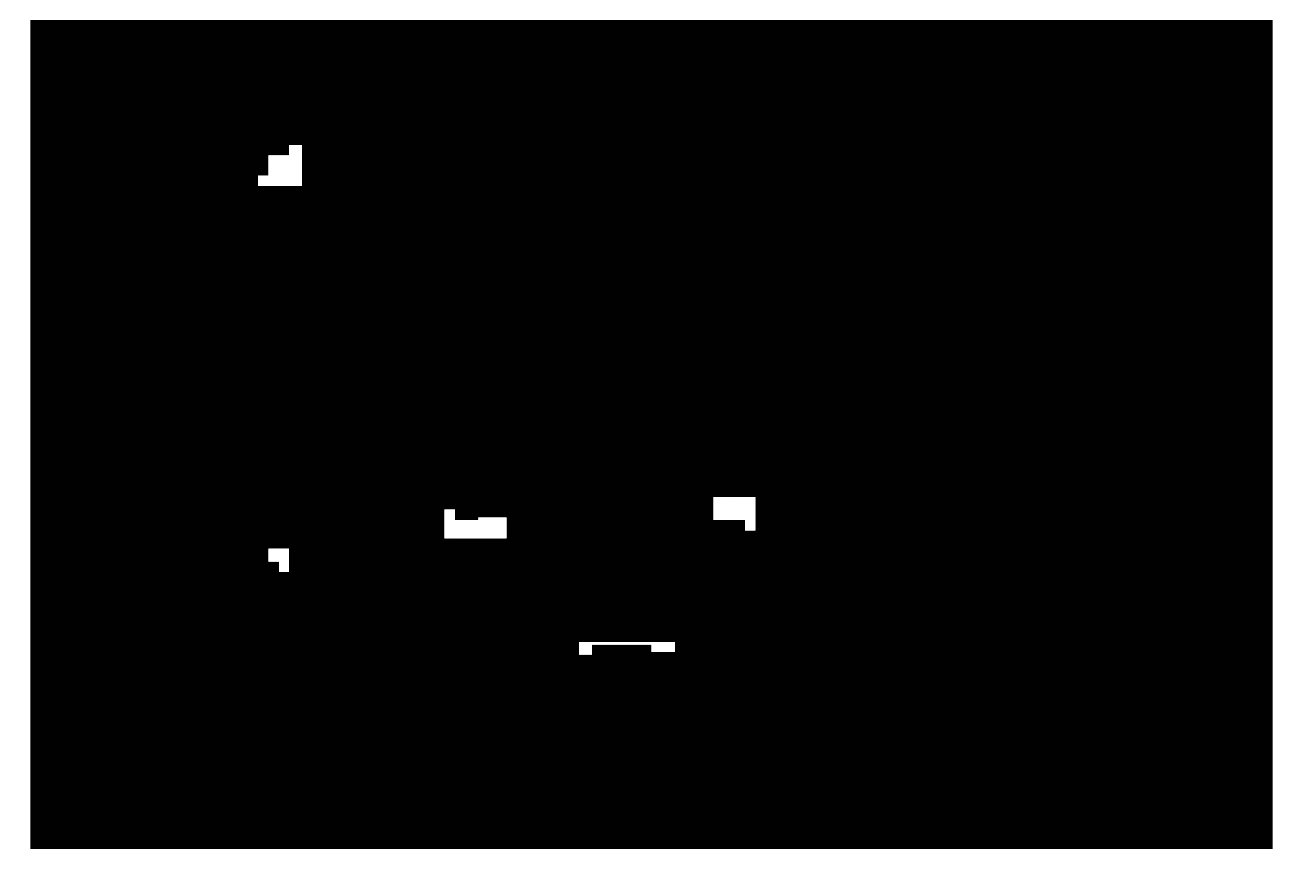}}
            \hfil            
        \subfloat[Result of UKF with 1 IMU and 2 GPS]{\includegraphics[width=\w\textwidth]{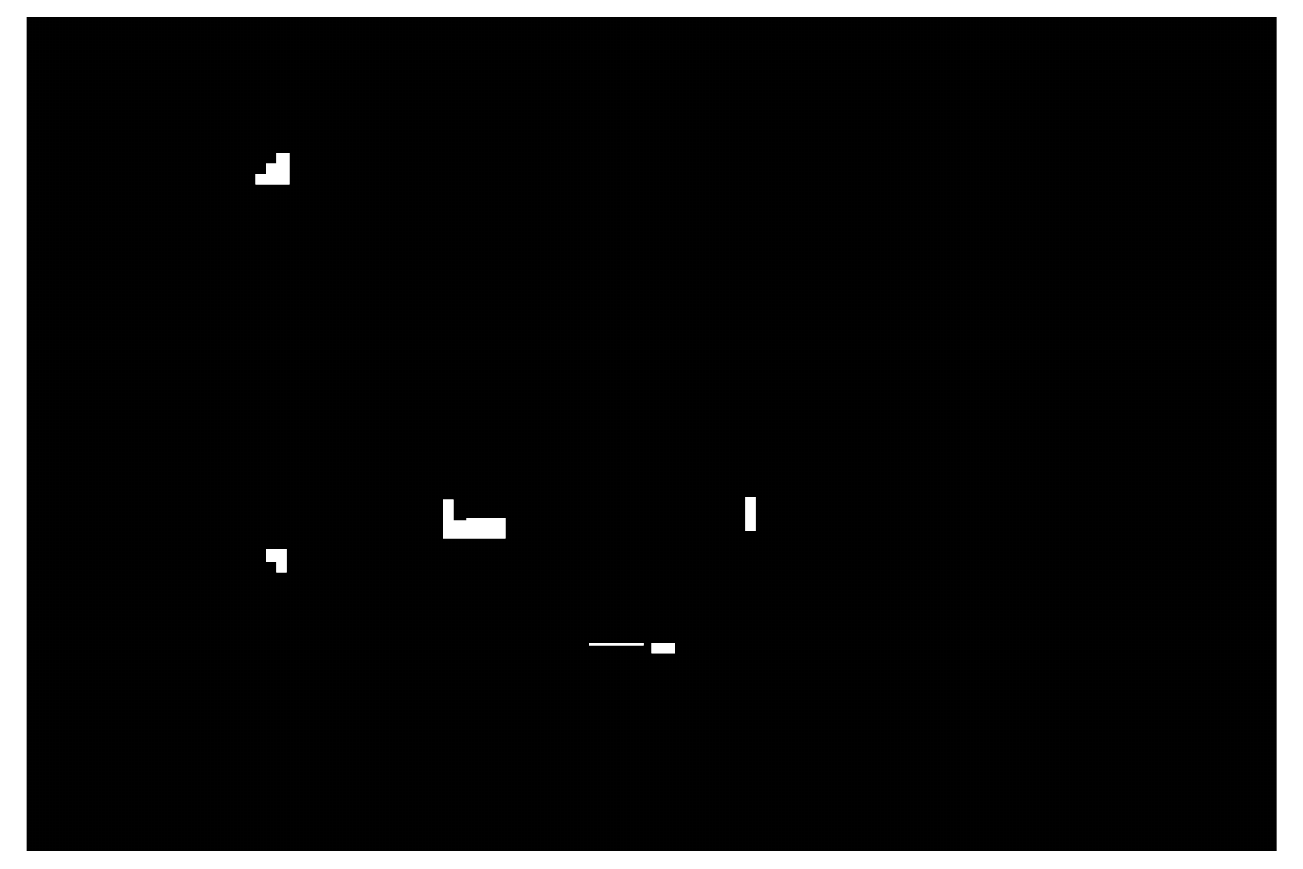}}
            \hfil            
        \caption{Difference between ground truth and the estimated map. Here we compare the predefined areas and the plotted path, where the white pixels are the wrong plotted grids. (a),(b),(c) are the Friction map results, and (d),(e),(f) are the road grade map results. UKF shows more accurate positioning capabilities than EKF, and with two GPS fused in the Kalman filter, the wrong located grid is less than just with one GPS signal.} 
        \label{fig:Comparison between the predefined areas and the plotted path}
\end{figure*}

As our ultimate goal is to create a map of the current working site in realtime based on localization technology so that corresponding optimization can then be achieved, the ground truth maps and estimated maps with different sensor configurations coupled with various algorithms are shown and compared in Fig. \ref{fig:After we drive the wheel loader around the simulation plane, the results of the plotter were created}.  Since we use a two-layer grid map, both rolling friction coefficient and road grade are recorded and used to create the estimated maps. In our study, the ground information are saved in corresponding grid after the wheel loader passes by and identify the ground information such as friction and slope through the specific algorithm, e.g. recursive least square. As aforementioned, to locate the mobile machine's position in a cost-efficient fashion, we adopt the configuration of the UKF with 1 IMU and 2 GPS. Also, in order to compare the performance of the selected configuration, we draw the results of the EKF with 1 IMU and 1 GPS and the UKF with 1 IMU and 1 GPS as the control groups. The mispredicted points are calculated as described 
in \eqref{eq:error_grid}, 

\begin{equation}
E_{r,s}= \sum\limits_{i=1, j=1}^{m, n} e_{i, j} =
\begin{cases} 
1,  & \text{if } \urcorner (\hat G_{i,j}= G_{i,j})  \cap  \urcorner (\hat G_{i,j}= NaN)  \\
0,  & \text{if }  (\hat G_{i,j}= G_{i,j}) \cap  \urcorner (\hat G_{i,j}= NaN) \\
0, &  \text{if } \hat G_{i,j}= NaN
\end{cases}
\label{eq:error_grid}
\end{equation}
where $ \hat G_{i,j} $ denotes the estimated grid map information, $ G_{i,j} $ is the ground truth map information, E is the accumulated number of errors, the subscript r and s denotes resistance and slope map, respectively. Obviously, the goal is to minimize the percent of mispredicted points versus total predicted points, described as \eqref{eq:goal},

\begin{equation}
    \text{min }  J_{r,s}= \frac{E_{r,s}}{T}
    \label{eq:goal}
\end{equation}
where T is the total estimated number, and J is the quantitative criteria to evaluate the accuracy of the plotted map.

The localization errors of each group are shown in Tab. \ref{tab:Localization errors of each approach during the plotting}, where we can see the necessity of the introduction of the second GPS sensors, and the adoption of unscented Kalman filter.

\begin{table}[h]
    \centering
    \caption{Localization errors of each approach during the plotting }
    \scalebox{0.8}{
    \begin{tabular}{|c|c|c|}
       \hline
       \multirow{2}*{\emph{\textbf{Group}}} &  \emph{\textbf{RMSE}} \textbf{(m)}& \multirow{2}*{\emph{\textbf{Net RMSE}}\textbf{(m)}} \\
        ~ & {\textbf{(x,y)}} & ~ \\
       \hline 1 (EKF 1 IMU 1 GPS) &  (2.584, 3.407)  & 4.4444  \\
       \hline 2 (UKF 1 IMU 1 GPS) &  (2.003, 2.314)  & 3.0603  \\
       \hline 3 (UKF 1 IMU 2 GPS) &  (1.336, 1.533)  & 2.0334  \\
       \hline 
    \end{tabular}}
    \label{tab:Localization errors of each approach during the plotting}
\end{table}

To evaluate the plotted maps' accuracy, we use Matlab to implement an algorithm to compare the predefined area and the plotted path the mobile machine traveled. Fig. \ref{fig:Comparison between the predefined areas and the plotted path} graphically illustrate the mispredicted grid point with white color, where the mispredicted points are generally distributed in the marginal zone. This is because even the localization technology makes some mistakes; the problem is unlikely to cause an error as long as the vehicle is not at the very edge of different zones, indicating the robustness of this mapping idea of the construction site.






After calculating the wrong located grids and all the plotted grids and according to the Eq. \eqref{eq:error_grid} and \eqref{eq:goal}:

\begin{itemize}
    \item [1.] For EKF with 1 IMU and 1 GPS: an error rate $J_r = 2.389\%$ of the ground resistances map and an error rate $J_s = 2.258\%$ of the slope map can be obtained.
    \item [2.] For UKF with 1 IMU and 1 GPS: an error rate $J_r = 1.482\%$ of the ground resistances map and an error rate $J_s = 1.682\%$ of the slope map can be obtained.
    \item [3.] For UKF with 1 IMU and 2 GPS: an error rate $J_r = 0.997\%$ of the ground resistances map and an error rate $J_s = 1.223\%$ of the slope map can be obtained.
\end{itemize}

\section{Conclusion}

In this study, we proposed an approach to creating a multi-layer map of the construction site in realtime so that the environmental information can be taken into account to improve vehicle efficiency and safety further or contribute to the path planning of mobile machines in the construction site. Considering the common phenomenon in reality mentioned by other researchers, such as noisy sensors and infrequent signal loss, we set up our simulation environment in Gazebo with a ROS package based on a real construction site. According to our tests in Gazebo by implementing a series of sensor configurations, we found that the configuration that 1 IMU and 2 GPS with encoder using UKF has the best for overall performance with respect to accuracy and cost. By comparing the estimated maps drawn by map plotter and the predefined maps, the errors are only 1.0\% and 1.2\% for road resistant force and grade, separately. Thus, we believe that the developed plotter can be used to save the road condition in realtime within a reasonable error range and encourage the engineer in construction machines to develop novel algorithms to further improve the holistic performance of machines based on our approach.

\subsection{Outlook}

In our research, we show the method to create a map with only one mobile machine. However, in a real working site, many mobile machines work simultaneously on the construction site, indicating the possibility of creating a map even faster if the machines can share the information.  Thus, we encourage the researchers to enable the cooperative map drawing approach by means of WIFI or 5G.

\bibliography{literature.bib}{}
\bibliographystyle{IEEEtran}



\end{document}